\documentclass{article}

\usepackage[margin=1in]{geometry}

\usepackage[colorlinks=true]{hyperref}

\usepackage{graphicx} 
\usepackage{url}
\usepackage{enumitem}
\usepackage{minted}
\usepackage{booktabs}
\usepackage{tabularray}
\usepackage{multirow}
\usepackage{rotating}
\usepackage{graphbox}

\usepackage{zref-clever}
\zcsetup{
    cap=true,
    nameinlink=false
}
\zcRefTypeSetup{algocf}{Name-sg=Algorithm,name-sg=algorithm}
\zcRefTypeSetup{algocfline}{Name-sg=Line,name-sg=line}

\usepackage{textcomp}
\newcommand{\mytexttilde}{\raisebox{0.5ex}{\texttildelow}}

\usepackage{footnote}
\makesavenoteenv{tabular}
\makesavenoteenv{table}

\usepackage{titling}
\predate{}
\postdate{}

\usepackage{tepper}

\usepackage[sortcites=true,natbib=true,doi=false,isbn=false,url=false]{biblatex}
\title{
    Individualized non-uniform quantization for vector search
}
\author{
    Mariano Tepper\\
    \texttt{mariano.tepper@ibm.com}\\
    IBM
    \and
    Ted Willke\\
    \texttt{ted.willke@ibm.com}\\
    IBM
}
\date{}

\begin{document}

\maketitle

\begin{abstract}
    Embedding vectors are widely used for representing unstructured data and searching through it for semantically similar items. However, the large size of these vectors, due to their high-dimensionality, creates problems for modern vector search techniques: retrieving large vectors from memory/storage is expensive and their footprint is costly. In this work, we present NVQ (non-uniform vector quantization), a new vector compression technique that is computationally and spatially efficient in the high-fidelity regime. The core in NVQ is to use novel parsimonious and computationally efficient nonlinearities for building non-uniform vector quantizers. Critically, these quantizers are \emph{individually} learned for each indexed vector. Our experimental results show that NVQ exhibits improved accuracy compared to the state of the art with a minimal computational cost.
\end{abstract}

\section{Introduction}

Modern embedding models are very adept at creating high-dimensional vectors whose spatial similarities reflect the semantic affinities between their inputs (e.g., images, audio, video, text, genomics, and computer code~\cite{devlin_bert_2019,radford_learning_2021,shvetsova_everything_2022,ji_dnabert_2021,li_competition-level_2022}). Many applications~\citep[e.g.,][]{blattmann_retrieval-augmented_2022,borgeaud_improving_2022,karpukhin_dense_2020,lian_lightrec_2020,grbovic_scalable_2016} have used this capability to find semantically relevant results in massive collections of vectors by retrieving the approximate nearest neighbors (ANN) of a given query vector. The most notable for its real-world deployment is retriever-augmented generation (RAG)~\citep{lewis_retrieval-augmented_2020}, using ANN to ground knowledge and to prevent LLMs from hallucinating.

Despite the outstanding progress in similarity search in the last few years~\citep[e.g.,][]{malkov_efficient_2020,fu_fast_2019,andre_quicker_2021,guo_accelerating_2020,subramanya_diskann_2019}, modern indices still struggle with high-dimensional vectors~\citep{aguerrebere_similarity_2023,tepper_leanvec_2024}.
Although moderate dimensionalities ($D \approx 128$) are handled accurately and with high performance~\citep{aguerrebere_similarity_2023}, efficiency suffers when dealing with the dimensionalities typically produced by embedding models ($D \approx 768$ and beyond)~\cite{tepper_leanvec_2024}. Next, we present two noteworthy examples of this behavior.

For in-memory indices (where the vectors and the index itself are held in main memory), graph-based search~\citep[e.g.,][]{arya_approximate_1993,malkov_efficient_2020,fu_fast_2019,subramanya_diskann_2019} is currently the state-of-the-art. Here, a directed graph, where each vertex corresponds to a database vector and edges represent neighbor-relationships between vectors, is efficiently traversed to find the approximate nearest neighbors in sub-linear time.
As we traverse the graph, vectors are fetched from memory in a random-like access pattern~\citep{aguerrebere_similarity_2023}, causing a memory bandwidth bottleneck. This high memory utilization drastically increases the memory latency \citep{srinivasan_cmp_2009} to access each vector, ultimately leading to suboptimal search performance. Even masterful software engineering~\citep{aguerrebere_similarity_2023} cannot hide the increased latency.

DiskANN~\citep{subramanya_diskann_2019} and SPANN~\citep{chen_spann_2021} are good representatives of the state of the art in the space of on-disk indices (where the vectors and the index itself are held in SSDs). DiskANN is a graph-based method that stores the high-dimensional vectors on disk next to the adjacency list of each node. With this layout, the same 4KB disk block can be used to store the list and the vector. However, this design does not hold for vectors with $1024$ dimensions or more since the vector itself occupies at least 4KB: here the vector payload does not come for free. SPANN is an inverted index that stores each vector in \mytexttilde$8$ clusters, with each cluster arranged contiguously on disk. Whereas this approach works for moderate-dimensional vectors, where each cluster can have a moderate size of 48KB~\citep{chen_spann_2021}, it creates an explosion of disk footprint and bandwidth utilization for high-dimensional vectors.

In addition, just storing large collections of high-dimensional vectors is a problem in itself; e.g., a relatively small collection of 10 million 1536-dimensional vectors occupies 57GB in single-precision format.
Existing vector compression techniques fall short as they are incompatible with random access patterns or do not provide sufficient accuracy~\citep{andre_quicker_2021,guo_accelerating_2020,aguerrebere_similarity_2023}.
New vector representations are needed that simultaneously present the following characteristics: (1) reduced footprint to alleviate memory/storage pressure and costs, (2) similarities must be preserved, and (3) similarity computations must be fast.

We tackle these problems by introducing \textbf{NVQ} (pronounced ``new vec''). NVQ is a non-uniform vector quantization method that is fitted to each vector \emph{individually} (in contrast to most vector quantizers that are fitted to the entire collection of vectors) to create more faithful representations (i.e., in terms of reconstruction error and similarities between vectors). Non-uniformity is embodied in NVQ through the novel use of parsimonious and computationally efficient nonlinearities, whose parameters are learned for each vector separately. This work presents the following detailed contributions:
\begin{itemize}[topsep=0ex,itemsep=0ex,parsep=0.ex,leftmargin=2.5ex]
    \item NVQ is the first method to learn a nonlinear quantizer individually for each vector. We also present an  optimization algorithm for this learning problem. We show that this formulation improves on the commonly used uniform quantization by a significant margin in terms of reconstruction error and vector search accuracy. Furthermore, we show that additional gains are obtained by dividing each vector into subvectors and individually applying NVQ to them. The code is available at \url{https://github.com/marianotepper/nuveq}.
    \item We propose the use of the Kumaraswamy distribution and of the logistic/logit nonlinearities for vector quantization. We also introduce a lightweight approximation of the logistic/logit nonlinearities that relies on not-quite transcendental functions. These nonlinearities are characterized by only two scalars and thus conduct to highly parsimonious and computationally efficient quantizers.
    \item We use NVQ in a production environment as part of an SSD-index for vector search. Here, NVQ reduces storage use by more than 3x versus 32-bit full-precision vectors with an impact of less than 0.01 on recall above 0.95. The code is available as part of JVector at \url{https://github.com/datastax/jvector/}.
\end{itemize}

The remainder of the work is structured as follows. We first review the related work in \zcref{sec:related}. In \zcref{sec:nvq} we present NVQ, the optimization problem behind it and the algorithm to solve it.
In \zcref{sec:nonlinearities}, we present three nonlinear functions that power NVQ, discussing the technical merits of each one in terms of expressiveness and efficiency.
We then present extensive experimental results in \zcref{sec:results} and provide concluding remarks in \zcref{sec:conclusions}.

\section{Related work}
\label{sec:related}

Research on approximate nearest neighbor (ANN) search has grown rapidly, motivated by the demands of applications like retrieval augmented generation, recommendation systems, and hybrid search, which involve increasingly larger datasets, higher dimensionality, and stringent recall and latency requirements.  ANN methods in the literature fall into different categories.  We focus the analysis on techniques suitable for large, high-dimensional datasets.  Tree-based approaches, such as ~\citep{bentley_multidimensional_1975, cayton_fast_2008, muja_scalable_2014, silpa-anan_optimised_2008}, struggle with the curse of dimensionality and do not generally scale well.  Hashing-based approaches, such as~\citep{indyk_approximate_1998, jafari_survey_2021, wang_survey_2018}, scale well but tend to suffer lower recall or use data-dependent hashing requiring full re-indexing or complex management pipelines as vectors are added. Graph-based approaches, such as~\citep{fu_fast_2019, malkov_efficient_2020, subramanya_diskann_2019, fu_high_2022, peng_efficient_2023}, typically provide better trade-offs between latency and recall than hashing and tree-based methods. Quantization-based methods, such as~\citep{jegou_product_2011, aguerrebere_similarity_2023, ge_optimized_2013, babenko_additive_2014, zhang_composite_2014, andre_quicker_2021, andre_cache_2015, ko_low-precision_2021, wang_deltapq_2020, gong_iterative_2013, paparrizos_fast_2022}, the category most relevant to our work, have become central due to their ability to compress large vector sets while supporting fast and accurate retrieval. Recent studies also propose hybrid algorithms, combining strengths from multiple categories to improve overall performance~\citep{subramanya_diskann_2019, jegou_product_2011, aguerrebere_similarity_2023, guo_accelerating_2020}. In this work, we present a new approach to quantization and evaluate it as a reranker in conjunction with a graph-based technique.  Comprehensive reviews and tutorials on ANN techniques can be found in recent surveys~\citep{pan_survey_2024, wang_comprehensive_2021, li_approximate_2020, aumuller_recent_2023, dobson_scaling_2023, wang_graph-_2023}.

Quantization reduces vector size by reducing the precision of each dimension, which saves memory and storage footprint while lowering bandwidth utilization.  Scalar is widely used due to its computational simplicity.  The most common form of scalar quantization is uniform quantization, in which a range of vector values are divided into equal intervals that are each assigned a single value.  This simplicity makes it fast but also introduces large reconstruction errors, which lowers recall.
Locally-adaptive quantization is a recent advancement in vector search that allows for the quantization parameters fitted individually at the vector level.  The first example of this for vector search is Locally-adaptive Vector Quantization~\citep{aguerrebere_similarity_2023}, where the data is first globally centered and then uniform scalar quantization is applied based on the minimum and maximum values within each vector.  NVQ is able to further reduce reconstruction error by dividing vectors into subvectors and fitting each more precisely through the use of simple nonlinear functions.

Vector quantization that takes into account the data distribution can lower reconstruction error for the same level of compression.  PQ~\citep{jegou_product_2011} is the most common distribution-aware technique.  The quantizer applies k-means clustering to M multi-dimensional subspaces of the input vector, encoding each subvector as the ID of the nearest centroid (e.g., an 8-bit unsigned integer).  PQ encoding can be trained using a small dataset sample.  It offers fast codebook generation, high compression ratios, and modest reconstruction error.  However, the resulting recall is generally too low to use it without reranking.  Additionally, as mentioned in~\citep{aguerrebere_similarity_2023}, PQ was designed primarily for inverted indices in which a precomputed table of partial similarities can be stored in transposed form.  This enables efficient SIMD computation of distances between a query and multiple database vectors simultaneously.  This transposition is not compatible with the random access patterns seen in graph-based search.  NVQ does not offer the compression ratios that PQ does, but it offers lower reconstruction error at levels of compression that are acceptable for storage and is easily vectorized for SIMD.  The Anisotropic Vector Quantization introduced in~\citep{guo_accelerating_2020} and used in ScaNN\footnote{\url{https://github.com/google-research/google-research/tree/master/scann}} is also a learned quantizer that builds on PQ, inheriting the same benefits and limitations.  

Aggressively compressing vectors to save memory often comes with a substantial reduction in recall.  A full-precision copy of the vectors may be retained in storage and used to boost recall in two-level quantization or with a final reranking step, as in~\citep{subramanya_diskann_2019, v_bang_2024}.  PQ codebooks and residuals are used in~\citep{xu_residual_2022} to boost recall, but the joint optimization of these levels is computationally prohibitive at large scale and the recall is lower than methods that utilize full-precision vectors.  SPANN in~\citep{chen_spann_2021} uses an inverted index, keeping only centroids in memory and storing multiple copies of vectors (associated with multiple clusters) on disk.  When used for storage-based reranking, NVQ significantly reduces the storage footprint and the amount of data transferred during I/O operations, benefitting these methods and others that store high-fidelity vectors on disk.

\section{Non-uniform vector quantization}
\label{sec:nvq}

A quantizer takes continuous values in a domain $\set{D}$ and discretizes them by mapping them to values in the set $\{0, 1, 2, \dots, 2^\beta - 1\}$ of natural numbers, where $\beta$ is the number of bits of the representation.
Formally, the $\beta$-bit quantizer $Q: \set{D} \to \set{B}_{\beta}$ and the $\beta$-bit dequantizer $Q^{-1} : \set{B}_{\beta} \to \set{D}$ are defined as
\begin{align}
    Q(x; h, \theta, \beta) &\defeq \left\lfloor (2^\beta - 1) h(x; \theta)  + 1/2 \right\rfloor
    \label{eq:quantizer}
    \\
    Q^{-1} (y; h, \theta, \beta) &\defeq h^{-1} \left( (2^\beta - 1)^{-1} y ; \theta \right) ,
    \label{eq:dequantizer}
\end{align}
where $h: \set{D} \to [0, 1]$ is an invertible nonlinearity with parameters $\theta \in \set{C}$. Although $Q$ is not an invertible function due to the floor function, we use the notation $Q^{-1}$ to denote its ``lossy inverse'' (its inverse when $\beta \rightarrow \infty$).

Of critical importance in this setting is the parsimony of the parameters $\theta$. In one extreme, we can achieve lossless quantization if we store the histogram of the values in the vector. Of course, this would not be sensible since transmitting $\{ Q(x_i; h, \theta, \beta) \}$ would be more expensive than transmitting $\vect{x}$ directly. On the other extreme, a uniform quantizer uses $h$ as the identity function, i.e., $h(x) = x$, and $\theta = \varnothing$. The main question of this work is: can we improve upon the uniform quantization using a data-driven approach with a highly parsimonious parameter set? We answer this question in the affirmative, showing that consistent improvements can be achieved using nonlinearities with only two scalar parameters.
In fact, this is not only possible, but computationally very efficient.

Traditionally, non-uniform quantizers are constructed by using the empirical CDF (i.e., its cumulative histogram) of the data. Here, we would use $h$ as an invertible parametric curve fitted to the empirical CDF. With this approach, we would allocate more bits to the most common data values. However, this is a problem for encoding vectors in the context of vector search, where we are less interested in the fidelity of the vectors themselves. What truly matters is fidelity of the similarity between the vectors and the queries.

Without loss of generality, we illustrate this point with maximum inner product (MIP) search, where the similarity between a query vector $\vect{q}$ and a database vector $\vect{x}$ is given by their dot product $\langle \vect{q}, \vect{x} \rangle$ and we seek to recover the vectors in $\set{X}$ that are most similar to the query. Extending these concepts to other popular similarity measures, such as Euclidean distance or cosine similarity is straightforward. In MIP,
\begin{equation}
    \langle \vect{q}, \vect{x} \rangle
    \approx
    \langle \vect{q}, \widetilde{\vect{x}} \rangle ,
    \quad\text{where}\quad
    \begin{aligned}
        \vect{y} &= Q(\vect{x}; h, \theta, \beta) ,\\
        \widetilde{\vect{x}} &= Q^{-1} (\vect{y}; h, \theta, \beta) .    
    \end{aligned}
\end{equation}
Here, the encoder/decoder are applied entry-wise for some appropriate selection of $h, \theta, \beta$, producing a quantized version $\widetilde{\vect{x}}$ of $\vect{x}$. Clearly, the entries of $\vect{x}$ with larger magnitudes have a larger impact on the dot product than those close to zero. However, we observe that modern embedding models produce vectors whose values present bell-shaped empirical value distributions, centered approximately at zero. Allocating more bits to these common values would yield a small pay-off.

Assuming a flat prior on the angular distribution of the queries (the norm of the query has no effect in MIP), we have~\cite{guo_accelerating_2020} for a $d$-dimensional vector $\vect{x} = \transpose{[x_1, \cdots, x_d]}$,
\begin{equation}
    \expectationFromDist{(\langle \vect{q}, \vect{x} \rangle - \langle \vect{q}, \widetilde{\vect{x}} \rangle )^2}{\vect{q}}
    =
    \norm{ \vect{x} - \widetilde{\vect{x}} }{2}^2
    =
    \sum_{i=1}^{d} \left( x_i - \widetilde{x}_i \right)^2 .
\end{equation}
Thus, minimizing the reconstruction error is the best approach for learning quantizers without additional information about the queries.

Given a suitable nonlinearity $h$, we seek the best parameters $\theta$ for each $\vect{x} \in \set{X}$. For this, we solve
\begin{equation}
    \min_{\theta \in \set{C}} \ell_{h}(\theta) 
    \quad\text{where}\quad
    \ell_{h}(\theta) \defeq \sum_{i=1}^{d} \left( x_i - Q^{-1} \left( Q(x_i; h, \theta, \beta) ; h, \theta, \beta \right) \right)^2
    .
   \label{prob:nuveq_least_squares}
\end{equation}
Notice that we perform no dataset-level optimization. Each vector is treated individually, finding the parameters that minimize its own reconstruction error.

We found that more informative loss values and stopping conditions can be achieved by working with the improvement over the uniform quantization baseline instead of the raw reconstruction error. That is, we solve
\begin{equation}
    \max_{\theta \in \set{C}}
    \frac{
        \ell_{\textsc{unif}}(\varnothing)
    }{
        \ell_{h}(\theta)
    }
    ,
   \label{prob:nuveq}
\end{equation}
where $\ell_{\textsc{unif}}(\varnothing)$ denotes the reconstruction error obtained with the parameterless uniform quantization. With this formulation, a value higher or lower than 1 indicates an improvement or decline in the quality of our quantization, respectively. It is important to note that Problem~\zcref[noname]{prob:nuveq} is $|\theta|$-dimensional. Its runtime depends on $d$ for a $d$-dimensional vector $\vect{x}$ but not its parameter space.

Solving Problem~\zcref[noname]{prob:nuveq} individually for each vector in the database has multiple benefits. First, it can be carried seamlessly when inserting each vector in the index. Second, it is completely robust to distribution shifts in the data that may occur over time~\cite{aguerrebere_locally-adaptive_2024,baranchuk_dedrift_2023}. Finally, and maybe more importantly, it enables ``overfitting'' the quantizer to each vector without downsides.

\subsection{The optimization}
\label{sec:optimization}

\begin{algorithm2e}[t]
    \small
    
    \KwIn{$f : \Real_+^m \to \Real$, $\vect{\mu}_{\text{init}} \in  \Real_+^m$, $\sigma_{\text{init}} \in  \Real_+^m$, $\eta_{\vect{\mu}} \in \Real_+$, $\eta_{\sigma} \in \Real_+$, $T \in \Nat_1$.}
    \KwOut{The solution $\vect{\mu}$.}
    \DontPrintSemicolon

    $\vect{\mu} \gets \vect{\mu}_{\text{init}}$\;
    
    \Repeat{stopping condition is met}{
        \For {$k = 1, \dots, T$}{
            draw sample $\vect{s}_k \sim \mathcal{N}(0, \mat{I})$\;
            $\vect{z}_k \gets \operatorname{project}_{\set{C}} \left( \vect{\mu} + \sigma \odot \vect{s}_k \right)$ \tcp*[h]{$\odot$ represents the Hadamard product}\;
            \label{line:nes_projection}
        }
        compute utilities $u_k$ for each $\vect{z}_k$ with respect to $f(\vect{z}_k)$ \citep[section 3.1]{wierstra_natural_2014}\;
    
        compute natural gradients
        $\left\{
        \begin{aligned}
            \nabla_{\vect{\mu}} J &\gets \textstyle \sum_{k=1}^{T} u_k \cdot \vect{s}_k
            \\
            \nabla_{\sigma} J &\gets \textstyle \sum_{k=1}^{T} u_k \cdot (\vect{s}_k^2 - 1)
        \end{aligned}
        \right.$
        
        update parameters
        $\left\{
        \begin{aligned}
            \vect{\mu} &\gets \max \left\{ \vect{\mu} + \eta_{\vect{\mu}} \sigma \cdot \nabla_{\vect{\mu}} J
             , \vect{0} \right\} \\
            \sigma &\gets \sigma \exp(\eta_{\sigma} / 2 \cdot \nabla_{\sigma} J )
        \end{aligned}
        \right.$
    }
    
    \caption{Separable NES with constraints}
    \label{algo:SNES}
\end{algorithm2e}

The nonlinearities $h$ and $h^{-1}$ make Problem~\zcref[noname]{prob:nuveq} non-convex.
Moreover, the objective function $\ell_{h}$ is discontinuous due to the floor function. This technical difficulty is commonly bypassed by using a straight-through estimator \citep{bengio_estimating_2013,vali_nsvq_2022}. In this work, we follow a gradient-free optimization approach, as described next.

\begin{figure}[p]
    \centering
    \begin{tblr}{
        colspec = {X[c,r,1]X[c,h,10]X[c,h,10]},
    }
        \begin{sideways}
            \small Kumaraswamy
        \end{sideways} &
        \includegraphics[width=\linewidth,align=c]{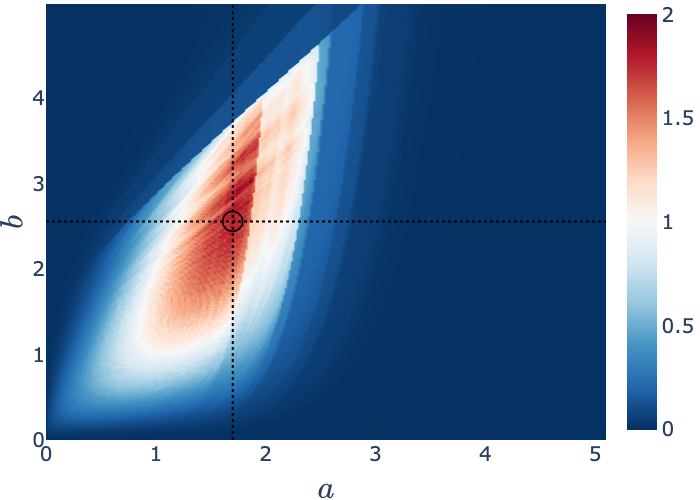} &
        \includegraphics[width=\linewidth,align=c]{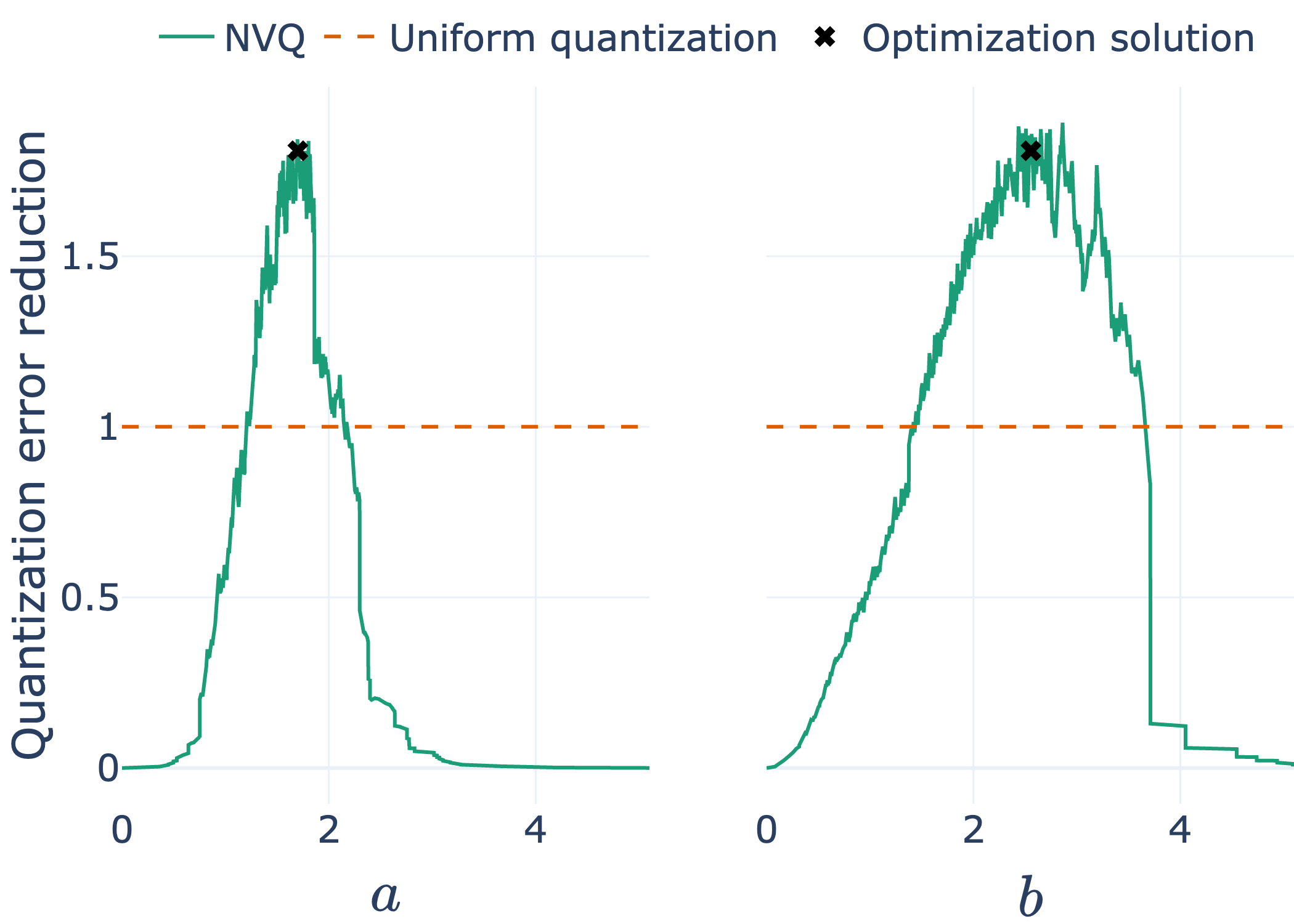} \\

        \begin{sideways}
            \small Log-Log
        \end{sideways} &
        \includegraphics[width=\linewidth,align=c]{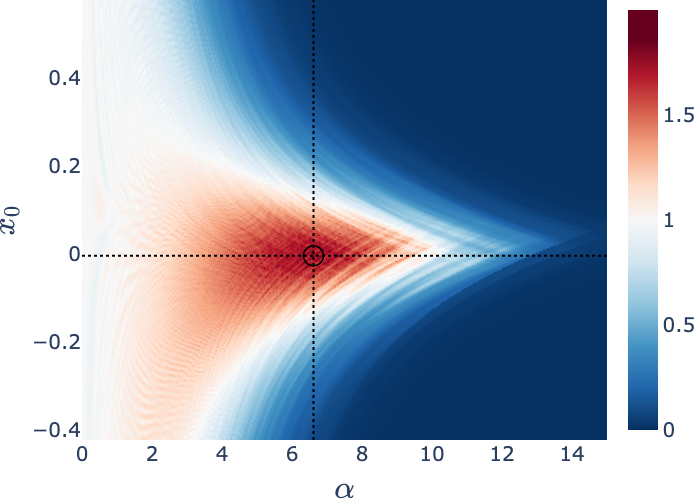} &
        \includegraphics[width=\linewidth,align=c]{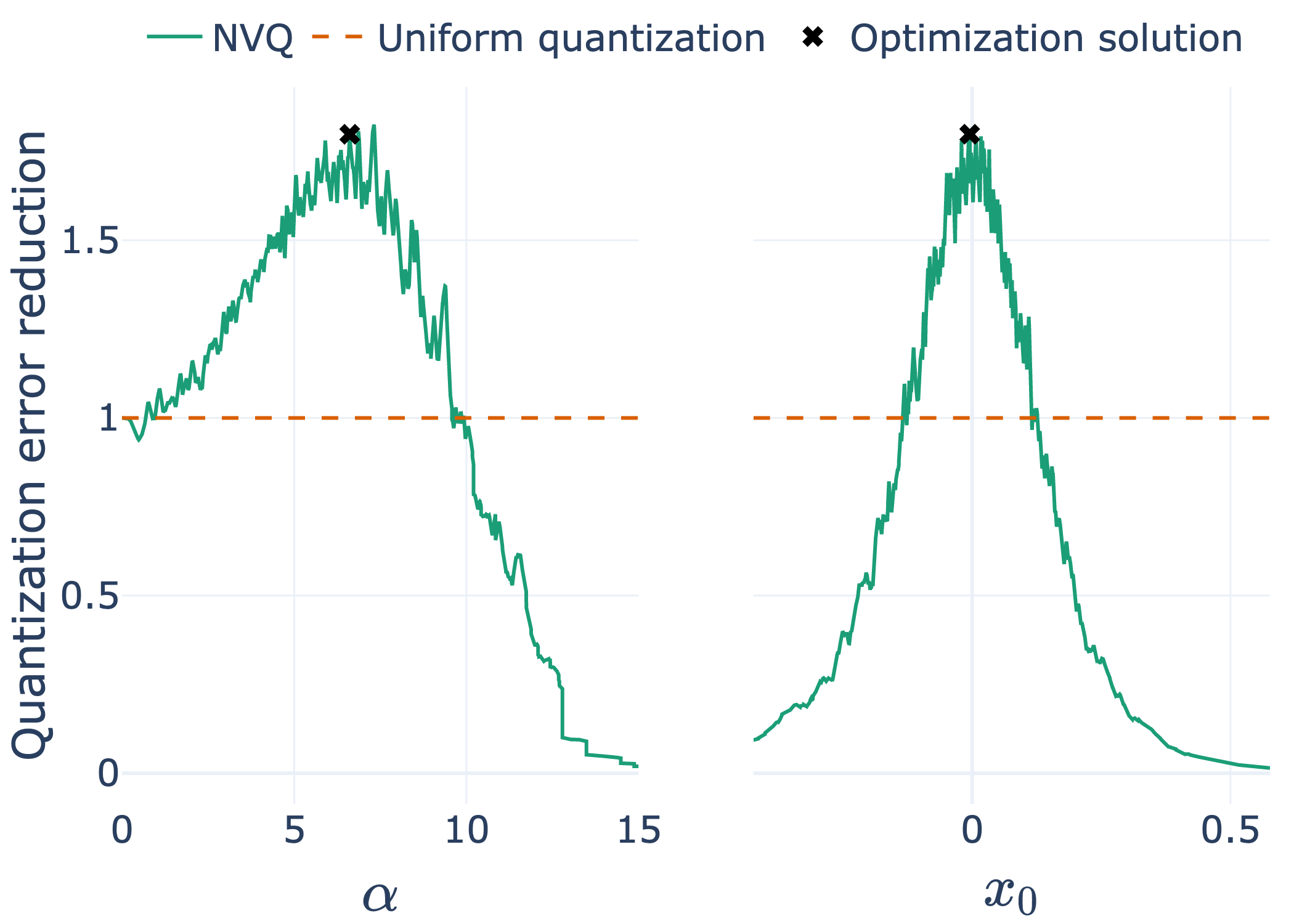} \\

        \begin{sideways}
            \small NQT
        \end{sideways} &
        \includegraphics[width=\linewidth,align=c]{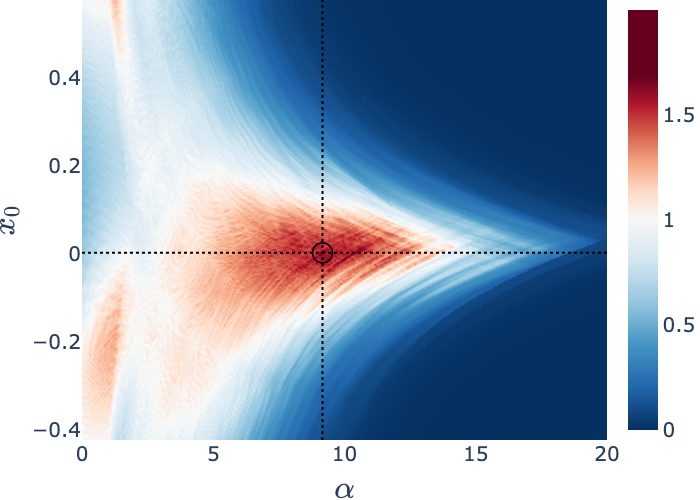} &
        \includegraphics[width=\linewidth,align=c]{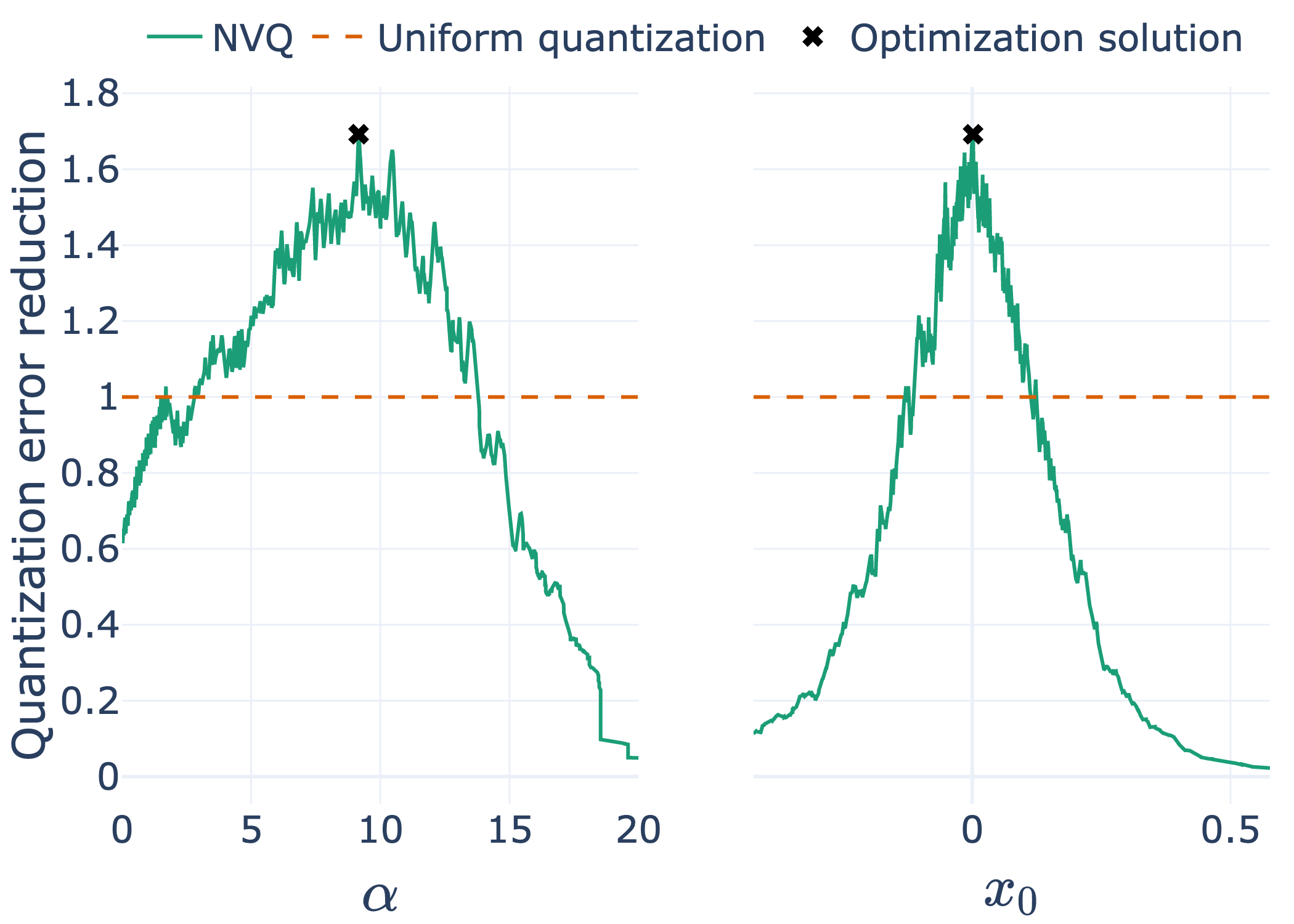} \\
    \end{tblr}
    
    \caption{\zcref{algo:SNES} finds a good maximum of Problem~\zcref[noname]{prob:nuveq} using the nonlinearities presented in \zcref{sec:nonlinearities} for $\beta=8$ bits (similar results for $\beta=4$ and a different dataset are provided in \zcref{fig:single_vector_ada002-100k-4bits,fig:single_vector_openai-8bits,fig:single_vector_openai-4bits} of the appendix). \textbf{Left:} the landscape of the objective function in Problem~\zcref[noname]{prob:nuveq}, which is the ratio of the MSE improvement over the uniform quantization (a value of 1 means parity, higher is better), as a function of the nonlinearity parameters for one vector from ada002-100k (see \zcref{tab:datasets}). The solution found by \zcref{algo:SNES} is marked by a black circle. \textbf{Right:} two cross cuts taken at the values corresponding to the found solution.}
    \label{fig:single_vector_ada002-100k-8bits}
\end{figure}

With the nonlinearities introduced in \zcref{sec:nonlinearities}, Problem~\zcref[noname]{prob:nuveq} is two-dimensional and we observe in \zcref{fig:single_vector_ada002-100k-8bits} that its landscape has a relatively large basin. Thus, it is amenable to gradient-free optimization techniques.\footnote{Alternatively, the two-dimensional nature of the problem allows to find a solution using grid search.} In this work, we optimize Problem~\zcref[noname]{prob:nuveq} using separable natural evolution strategies \citep{wierstra_natural_2014}, which estimate a proxy of the gradient by computing the objective function at stochastic samples around the current estimate of the solution. 
In \zcref{algo:SNES} we modify Algorithm 6 in \citep{wierstra_natural_2014} to account for the constraint $\theta \in \set{C}$ in Problem~\zcref[noname]{prob:nuveq}. The modification consists of adding a projection step in \zcref{line:nes_projection}. This step takes different forms depending on the nonlinearity, as described in \zcref{sec:nonlinearities}.

In \zcref{algo:SNES}, we use $f(\cdot) = \ell_{\textsc{unif}}(\varnothing) / \ell_{h}(\cdot)$.
For the hyperparameters, we approximately adopt the suggestions in~\citep{wierstra_natural_2014}: the number of stochastic samples is $T = 2 (4 + \lfloor 3 \log |\theta| \rfloor ))$ and the learning rates are $\eta_{\vect{\mu}} = 1$ and $\eta_{\sigma} = (9 + 3 \log |\theta| ) / (5 m \sqrt{|\theta|})$. The initial values $\vect{\mu}_{\text{init}}$ and $\sigma_{\text{init}}$ are specific to each nonlinearity, as described in \zcref{sec:nonlinearities}.
Unless specified, we use the stopping condition $\left| \vect{\mu}^{(t)} - \vect{\mu}^{(t-1)} \right| < 10^{-4}$, where $\vect{\mu}^{(t)}$ is the value of $\vect{\mu}$ at the $t$-th iteration. We also impose a minimum of 10 iterations.

\subsection{Quantizing subvectors}
\label{sec:subvectors}

Inspired by PQ~\cite{jegou_product_2011}, we consider dividing a vector into subvectors. We divide its $d$ dimensions at random into $m$ sets, forming $m$ subvectors with $d/m$ dimensions each, i.e., $\vect{x} = [\vect{x}^{(1)}, \vect{x}^{(2)}, \dots, \vect{x}^{(m)}]$. In this case, we optimize Problem~\zcref[noname]{prob:nuveq} for each subvector individually and we have $\theta = \{ \theta^{(1)}, \theta^{(2)}, \dots, \theta^{(m)} \}$. With modern embedding models, using $m \in \{2, 4, 8\}$ ensures that $d$ is a multiple of $m$.

The payload of quantizing subvectors is obviously higher than that of quantizing entire vectors. However, this cost is relatively small when compared to the size of the vector themselves. For 1024-dimensional vectors, it the vector values occupy 1024 bytes for $\beta=8$. As we describe in the next section, for $m \in \{2, 4, 8\}$, we have $|\theta|= \{8, 16, 32 \}$ parameters ($|\theta^{(j)}| = 4$ for $j = 1,\cdots, m$), which occupy 32B, 64B, and 128B, respectively. We finally point out that it is often desirable to store the vectors in having a cache-aligned memory layout~\citep{aguerrebere_similarity_2023}. From this point of view, the parameters can be stored in the ``free space'' used to produce this alignment.

\section{The nonlinearities}
\label{sec:nonlinearities}

We now introduce the nonlinear functions used in Problem~\zcref[noname]{prob:nuveq}. These functions, which are used for vector quantization for the first time in this work, are characterized by their parsimony, requiring only two scalar parameters, and their computational efficiency. We present three alternatives that strike different trade-offs between expressiveness and computational efficiency, as shown in \zcref{tab:computational_costs}.

Each compressed vector needs two additional values in its payload, $x_{\text{min}} = \min_i x_i$ and $x_{\text{max}} = \max_i x_i$. These two values are not optimized as part of Problem~\zcref[noname]{prob:nuveq}, they are used in different ways to normalize the each set of nonlinearities as described in each subsection.

\begin{table}[t]
    \caption{Expressiveness and computational cost of the three nonlinearities used in this work. We measure the cost in terms of fundamental functions and main computer instructions (for simplicity, we do not count additions, subtractions, or bit operations). While the Kumaraswamy distribution is the most expressive nonlinearity (see \zcref{fig:kumaraswamy_parameter_examples}), its ``heavy'' use of fundamental functions makes it computationally demanding. The other two nonlinearities are less expressive (but sufficiently so for vectors from embedding models) and computationally nimbler. Implementation details in \zcref{sec:fast_exponentiation,sec:nqt_code}.}
    \label{tab:computational_costs}

    \centering
    \small
    \begin{tblr}{
            colspec = {llccc},
            rowspec = {|c|c|cc|ccc|cc|ccc|},
            rowsep=1pt,
        }
             && Kumaraswamy & Logistic/Logit & NQT Logistic/Logit \\
        \SetCell[c=2]{c} Expressiveness && multiple distributions & bell-shaped & bell-shaped \\
        
        \SetCell[r=5]{c,m} {Encoding \\ cost}
        & $\exp$ & 2           & 1              & 0 \\
        & $\log$ & 2           & 0              & 0 \\
        & FMA    & 18          & 6              & 2 \\
        & MUL    & 8           & 2              & 0 \\
        & DIV    & 0           & 1              & 1 \\

        \SetCell[r=5]{c,m} {Decoding \\ cost}
        & $\exp$ & 2           & 0              & 0 \\
        & $\log$ & 2           & 1              & 0 \\
        & FMA & 18          & 7              & 2 \\
        & MUL & 8           & 2              & 0 \\
        & DIV & 0           & 1              & 1 \\
    \end{tblr}
\end{table}

\subsection{The generalist: The Kumaraswamy distribution}

\begin{figure}[t]
    \centering
    \includegraphics[width=0.6\linewidth]{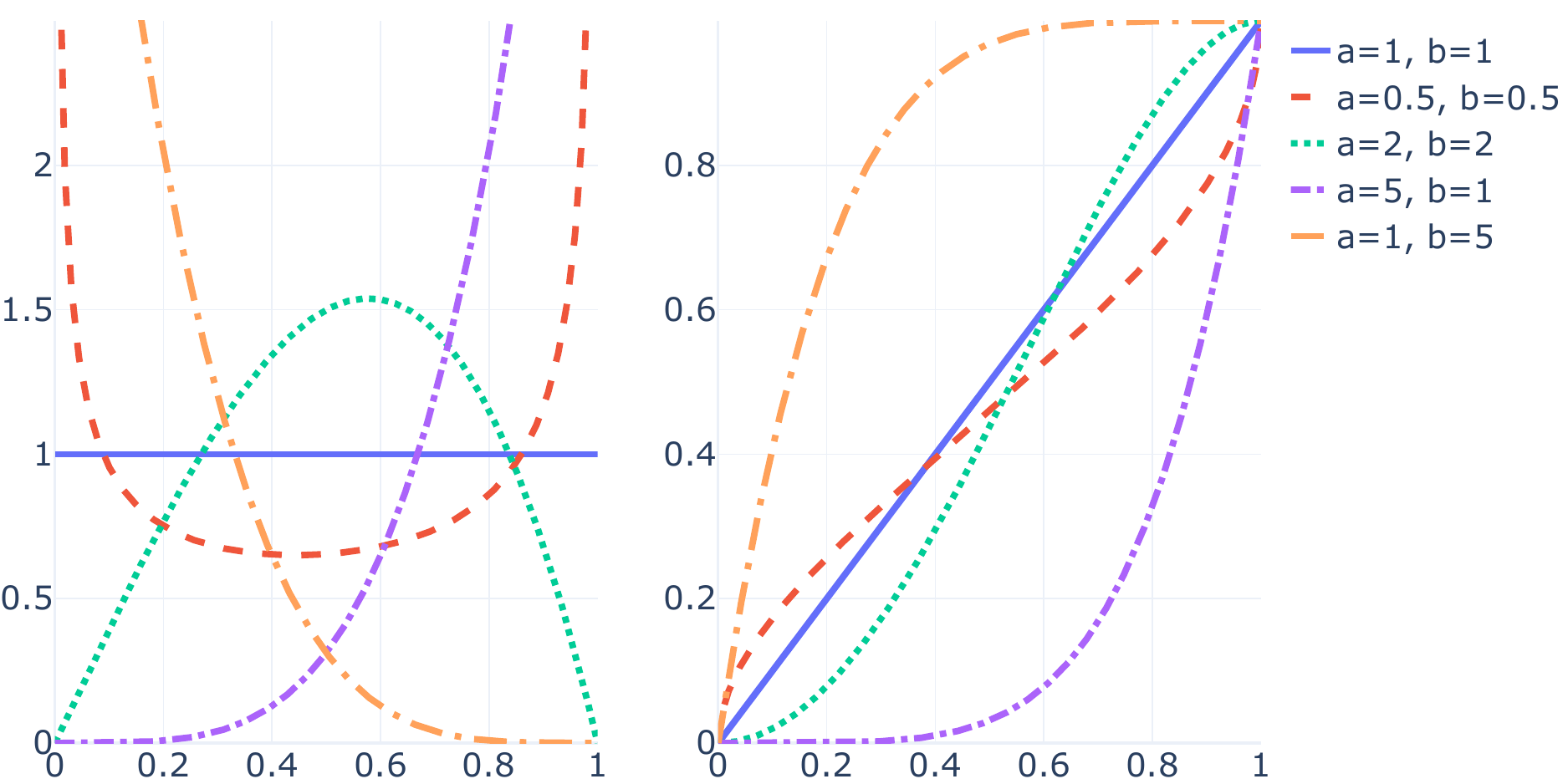}
    
    \caption{The PFD (left) and CDF (right) of the Kumaraswamy distribution, defined in \zcref{eq:kumaraswamy_cdf}, for different values of $a$ and $b$. The Kumaraswamy distribution allows to represent sample distributions with a variety of shapes.}
    \label{fig:kumaraswamy_parameter_examples}
\end{figure}

The wide use of the beta distribution is a testament of its expressiveness to model different phenomena in the interval $[0, 1]$. However, it is computationally expensive as its partition function if given by the beta function, that lacks a closed form. The Kumaraswamy distribution is similar to the beta distribution in expressiveness, see \zcref{fig:kumaraswamy_parameter_examples}, but much more computationally amenable since its probability density, cumulative distribution, and quantile functions can be expressed in closed form~\citep{kumaraswamy_generalized_1980,jones_kumaraswamys_2009,wasserman_stabilizing_2024}.
Its cumulative distribution and its quantile functions are respectively given by
\begin{align}
    F_{\textsc{ks}}(x; a, b) &\defeq 1 - \left(1 - x^a \right)^b ,
    \label{eq:kumaraswamy_cdf}
    \\
    F_{\textsc{ks}}^{-1}(y; a, b) &\defeq \left(1 - \left( 1 - y \right)^{1/b} \right)^{1/a} ,
    \label{eq:kumaraswamy_icdf}
\end{align}
where $x \in [0, 1]$ and $a, b \in \Real_+$. The Kumaraswamy distribution is traditionally defined in the open interval $(0, 1)$. Since $F$ is well defined for $x=0$ and $x=1$, we extend the range to the closed interval.
To operate in $[0, 1]$, we normalize the values $x_i$ in $\vect{x}$ as $(x_i - x_{\text{min}}) / (x_{\text{max}} - x_{\text{min}})$, where $x_{\text{min}} = \min_i x_i$ and $x_{\text{max}} = \max_i x_i$.

We propose to solve Problem~\zcref[noname]{prob:nuveq} with the parametrization
\begin{align}
    \theta &\defeq [a, b] ,
    &
    h &\defeq F_{\textsc{ks}} ,
    &
    h^{-1} &\defeq F_{\textsc{ks}}^{-1} .
\end{align}
Using this parametrization in \zcref{{eq:quantizer},eq:dequantizer} makes the pair $Q, Q^{-1}$ highly adaptable with only two parameters (the scalars $a, b$) and involves a closed-form computation.

For the Kumaraswamy nonlinearity, $a, b \in \Real_+$. The projection in \zcref{line:nes_projection} of \zcref{algo:SNES} takes the form
\begin{align}
    \operatorname{project}_{\set{C}} \left( a \right)
    &\defeq
    \max \left\{ a , 0 \right\} ,
    &
    \operatorname{project}_{\set{C}} \left( b \right) ,
    &\defeq
    \max \left\{ b , 0 \right\} ,
\end{align}
and we use the initial values $\vect{\mu}_{\text{init}} = [1, 1]$ and $\sigma_{\text{init}} = [1, 1]$.

\paragraph{Efficiency:}
We implement the power operator using the identity $x^c = \exp{ \left( c \log{x} \right) }$. We note that numerically stable formulations have been recently proposed~\citep{wasserman_stabilizing_2024} but we did not encounter such problems in our gradient-free optimization algorithm.
The logarithm and exponential functions can be implemented with sufficient accuracy using minimax polynomial approximations.
Additional implementation details are provided in \zcref{sec:fast_exponentiation}.
These highly flexible nonlinearities are relatively fast to compute, as observed in \zcref{tab:computational_costs}.

\subsection{The specialist: The scaled logistic and scaled logit}
\label{sec:logistic}

The Kumaraswamy distribution imbues NVQ with a high degree of flexibility. However, the computation of powers is an expensive operation that does not commonly have native support in CPUs or GPUs. Using exponentials and logarithms, as described in \zcref{sec:fast_exponentiation}, alleviates but does not eliminate this problem.

\begin{figure}
    \centering
    \includegraphics[width=0.3\linewidth]{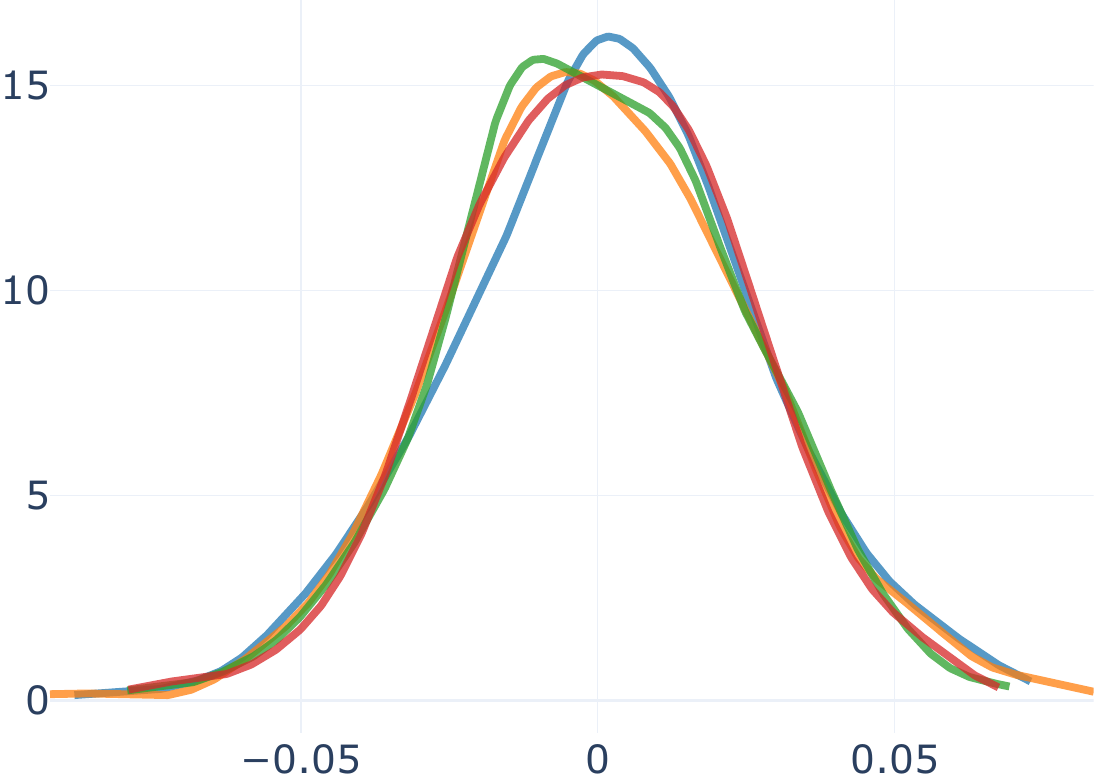}
    \includegraphics[width=0.3\linewidth]{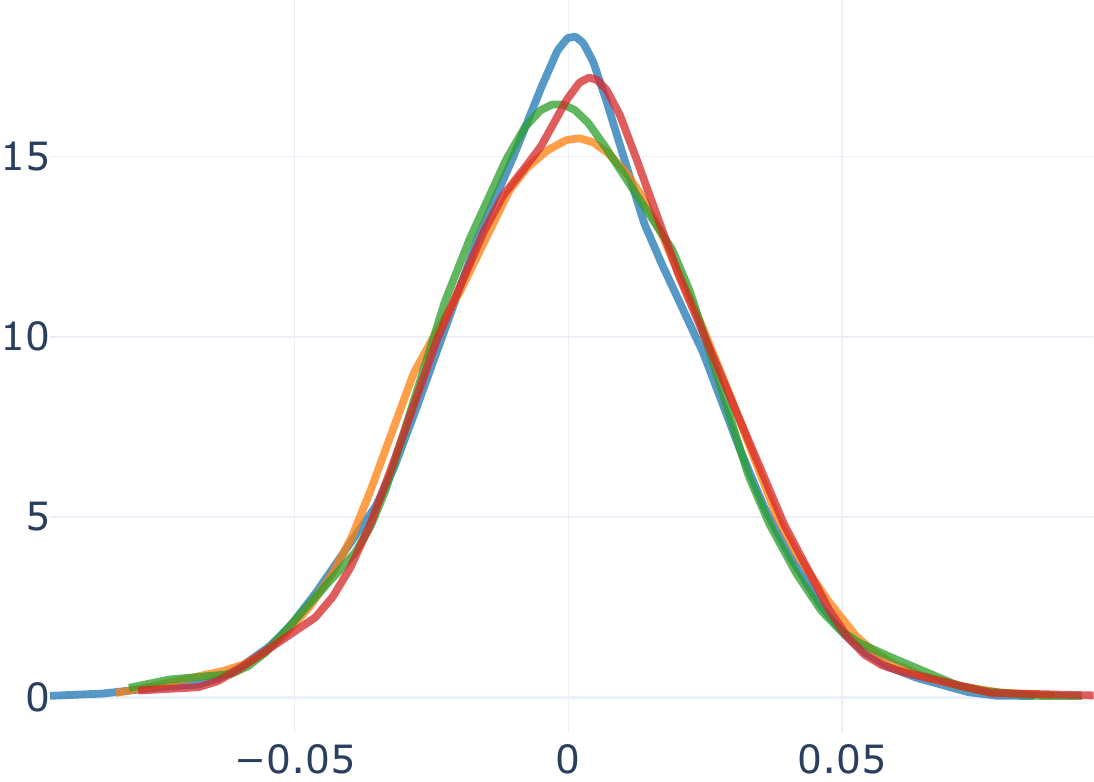}
    \caption{The kernel density estimate of PDF of the values from four different vectors in gecko-100k (left) and in openai-v3-100k (right). The sample PDFs are bell-shaped and, although they are all generally similar, they are not exactly the same. We can improve the quantization by tuning it for each vector individually.}
    \label{fig:histograms}
\end{figure}

By examining the empirical distribution of the values in vectors produced by modern embedding models, we see in \zcref{fig:histograms} that they are bell shaped. Thus, we are not fully exploiting the Kumaraswamy distribution's flexibility: an optimal solution to Problem~\zcref[noname]{prob:nuveq} for these vectors would never include parameters that generate exponential-like distributions even if the Kumaraswamy distribution supports this choice.\footnote{The use of the Kumaraswamy still holds value as future embedding models may produce vectors with other distributions.} Therefore, seeking computational efficiency, we turn our attention to nonlinearities that are specifically well suited to handle bell-shaped distributions.

\begin{figure}[t]
    \centering
    \includegraphics[width=0.75\linewidth]{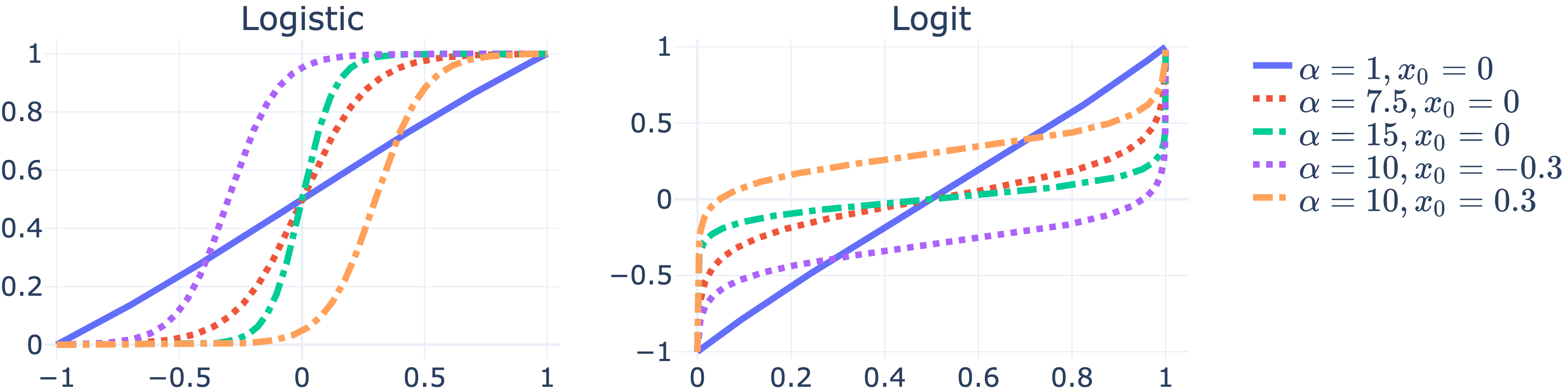}
    \caption{The scaled logistic (left) and logit (right) functions, defined in \zcref{eq:logistic_scaled,eq:logit_scaled}, for different values of $\alpha$ and $x_0$ (in these examples $x_{\text{min}} = -1$ and $x_{\text{max}} = 1$). These nonlinearities allow to approximate bell-shaped distributions.}
    \label{fig:logistic_parameter_examples}
\end{figure}

The standard logistic and logit functions on $\Real \to (0, 1)$ and $(0, 1) \to \Real$, respectively, are defined as
\begin{align}
    \operatorname{logistic} (x; \alpha, x_0) &\defeq \left( 1 + \exp(-\alpha (x - x_0)) \right)^{-1} ,
    \label{eq:logistic}
    \\
    \operatorname{logit} (y; \alpha, x_0) &\defeq \alpha^{-1} \log \left( \frac{y}{1 - y} \right) + x_0 .
    \label{eq:logit}
\end{align}
Let $x_{\text{min}} = \min_i x_i$ and $x_{\text{max}} = \max_i x_i$. We are interested in quantizing values in $[x_{\text{min}}, x_{\text{max}}]$. For this, we define scaled versions of these functions, i.e., $\operatorname{logistic}_{\textsc{scaled}} : [x_{\text{min}}, x_{\text{max}}] \to [0, 1]$ and $\operatorname{logit}_{\textsc{scaled}} : [0, 1] \to [x_{\text{min}}, x_{\text{max}}]$, as follows
\begin{align}
    \operatorname{logistic}_{\textsc{scaled}}(x; \alpha, x_0)
    &\defeq
    \frac{\operatorname{logistic} (\delta^{-1} x; \alpha, x_0) - \operatorname{logistic} (\delta^{-1} x_{\text{min}}; \alpha, x_0)}{\Delta} ,
    \label{eq:logistic_scaled}
    \\
    \operatorname{logit}_{\textsc{scaled}} (y; \alpha, x_0)
    &\defeq
    \delta
    \operatorname{logit} \big( \Delta y + \operatorname{logistic} (\delta^{-1} x_{\text{min}}; \alpha, x_0) \big) ,
    \label{eq:logit_scaled}
\end{align}
where $\delta = x_{\text{max}} - x_{\text{min}}$ and $\Delta = \operatorname{logistic} (x_{\text{max}}; \alpha, x_0) - \operatorname{logistic} (x_{\text{min}}; \alpha, x_0)$.
As depicted in \zcref{fig:logistic_parameter_examples}, these scaled functions have the domain and image of interest. Scaling of the input values $x$ by $\delta^{-1}$ makes the parameters $\alpha$ and $x_0$ comparable across vectors by making them invariant to the specific domain $[x_{\text{min}}, x_{\text{max}}]$ of each vector (this scaling is equivalent to using the alternative parametrization $\widetilde{\alpha} = \delta^{-1} \alpha$ and $\widetilde{x}_0 = \delta x_0$).

We solve Problem~\zcref[noname]{prob:nuveq} with the parametrization
\begin{align}
    \theta &\defeq [\alpha, x_0] ,
    &
    h &\defeq \operatorname{logistic}_{\textsc{scaled}} ,
    &
    h^{-1} &\defeq \operatorname{logit}_{\textsc{scaled}} .
\end{align}
It provides enough adaptability to the pair the pair $Q, Q^{-1}$ and brings consistent improvements over a uniform quantization.

The computation is simpler than those involved when using the Kumaraswamy distribution: it only involves one exponential for $h$ and one logarithm for $h^{-1}$, while $h$ and $h^{-1}$ involve two exponentials and two logarithms with the Kumaraswamy distribution. Moreover, only two parameters are still sufficient to characterize these nonlinearities.

For the logistic/logit nonlinearities, $\alpha \in \Real_+$ and $x_0 \in [x_{\text{min}}, x_{\text{max}}]$. The projection in \zcref{line:nes_projection} of \zcref{algo:SNES} takes the form
\begin{align}
    \operatorname{project}_{\set{C}} \left( \alpha \right)
    &\defeq
    \max \left\{ \alpha , 10^{-6} \right\} ,
    &
    \operatorname{project}_{\set{C}} \left( x_0 \right) ,
    &\defeq
    \min \left\{ \max \left\{ x_0 , x_{\text{min}} \right\} , x_{\text{max}} \right\} ,
    \label{eq:project_logistic}
\end{align}
and we use the initial values $\vect{\mu}_{\text{init}} = [10, 0]$ and $\sigma_{\text{init}} = [2, 0.5]$.

\paragraph{Efficiency:}
The logistic and logit nonlinearities, as the Kumraswamy distribution, rely on fundamental functions (exponentials and logarithms) for their computation. However, their use is significantly reduced, as observed in \zcref{tab:computational_costs}, resulting in a significant acceleration.

\subsection{The speed demon: Not-quite transcendental nonlinearities}
\label{sec:nqt}

The logistic and logit functions provide a good fit for vectors commonly encountered in modern ANN and a good acceleration over the Kumaraswamy distribution. However, it still relies on fundamental functions (exponentials and logarithms), which, even if used more sparingly, are still computationally expensive. We now show an alternative that bypasses fundamental functions altogether by using tight approximations.

The traditional approach to accelerate the computation of fundamental functions is using a low-degree polynomial approximation, e.g., as in \zcref{sec:fast_exponentiation}. In our application, we do not necessarily need to rely on fundamental functions as we only care about reasonable non-linearities for Problem~\zcref[noname]{prob:nuveq}. Thus, we can create our own nonlinearity instead of tightly approximating a given function. As detailed next, this leads to a significant acceleration.

Our starting point for these new nonlinearities are the base-2 logistic and logit functions
\begin{align}
    \operatorname{logistic}_2 (x; \alpha, x_0) &\defeq \left( 1 + 2^{(-\alpha (x - x_0))} \right)^{-1} ,
    \label{eq:logistic_base2}
    \\
    \operatorname{logit}_2 (y; \alpha, x_0) &\defeq \alpha^{-1} \log_2 \left( \frac{y}{1 - y} \right) + x_0 .
    \label{eq:logit_base2}
\end{align}
Qualitatively, these functions behave exactly like their natural counterparts. However, the binary base is useful for dealing with floating-point numbers.
A positive floating-point number $z$ is internally represented as $z = m \cdot 2^p$, where $m \in [0.5, 1)$ is the mantissa and the integer $p$ is the exponent. Then, to implement $\operatorname{logit}_2$ we can use the identity 
\begin{equation}
    \log_2 (z) = \log_2 (m) + p .
\end{equation}
The not-quite transcendental (NQT) function~\citep{miller_not-quite_2022} is a piecewise-linear interpolation of $\log_2$~\cite{hall_generation_1970}, defined as
\begin{equation}
    \log_{\textsc{nqt}} (z) \defeq 2 (m - 1) + p .
\end{equation}
With these elements, we can introduce our new nonlinearities.

\begin{definition}
    Let $y \in (0, 1)$, $\alpha \in \Real_+$, and $x_0 \in \Real$. Let $m \cdot 2^p$ be the binary representation of $z = y / (1 - y)$. We define the NQT logit as
    \begin{equation}
        \operatorname{logit}_{\textsc{nqt}} (y; \alpha, x_0)
        \defeq
        \alpha^{-1} \left[ 2 (m - 1) + p \right] + x_0
        .
        \label{eq:logit_nqt}
    \end{equation}    
\end{definition}

The NQT logit is continuous, piecewise differentiable, monotonically increasing, and invertible. Using simple manipulations, we derive its inverse as defined next.

\begin{definition}
    Let $x \in \Real$, $\alpha \in \Real_+$, and $x_0 \in \Real$. The NQT logistic function, as
    \begin{equation}
        \operatorname{logistic}_{\textsc{nqt}} (x; \alpha, x_0)
        \defeq \frac{m \cdot 2^p}{m \cdot 2^p + 1}
        \quad\text{where}\quad
        \begin{aligned}
            p &= \left\lfloor \alpha (x - x_0) + 1 \right\rfloor ,
            \\
            m &= \frac{\alpha (x - x_0) - p}{2} + 1 .
        \end{aligned}
        \label{eq:logistic_nqt}
    \end{equation}
\end{definition}

For our NQT-based quantization, we define $\operatorname{logistic}_{\textsc{nqt}, \textsc{scaled}}$ and $\operatorname{logit}_{\textsc{nqt}, \textsc{scaled}}$, the scaled versions of $\operatorname{logistic}_{\textsc{nqt}}$ and $\operatorname{logit}_{\textsc{nqt}}$ exactly as in \zcref{eq:logistic_scaled,eq:logit_scaled}; we omit the equations for brevity. We solve Problem~\zcref[noname]{prob:nuveq} with the parametrization
\begin{align}
    \theta &\defeq [\alpha, x_0] ,
    &
    h &\defeq \operatorname{logistic}_{\textsc{nqt}, \textsc{scaled}} ,
    &
    h^{-1} &\defeq \operatorname{logit}_{\textsc{nqt}, \textsc{scaled}} .
\end{align}

\begin{figure}
    \centering
    \includegraphics[width=0.33\linewidth]{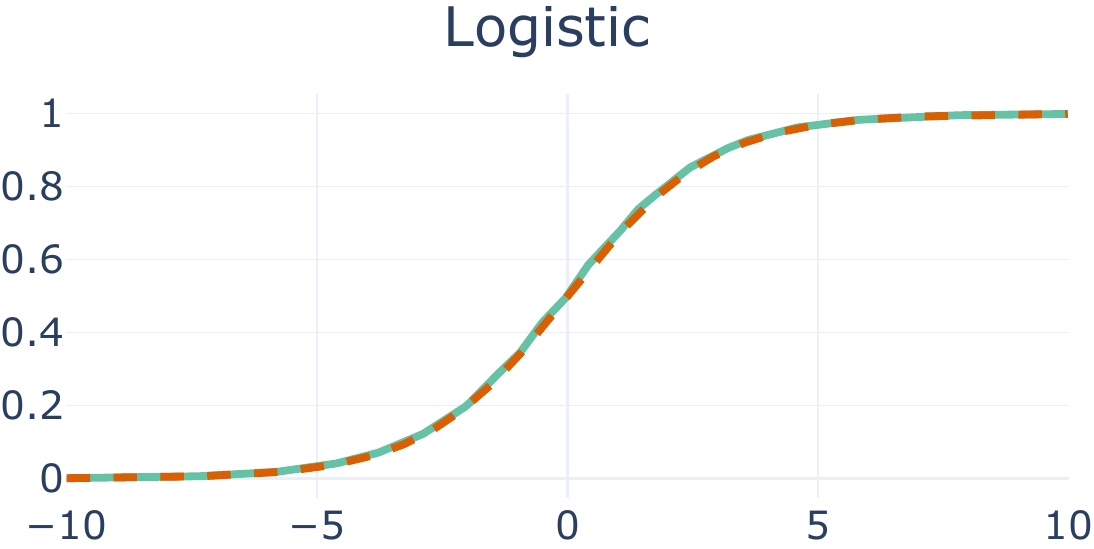}%
    \hfill%
    \includegraphics[width=0.33\linewidth]{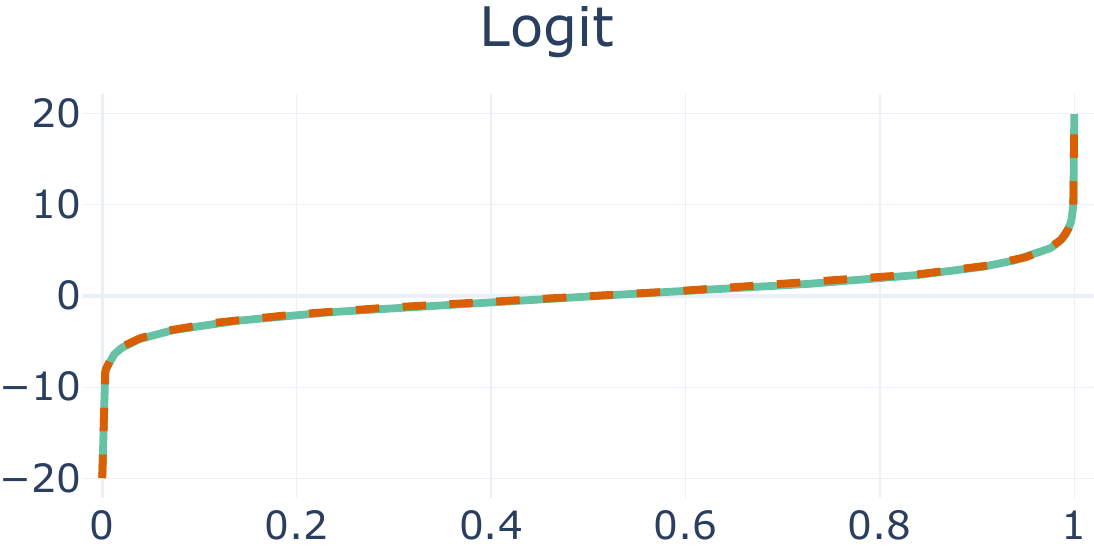}%
    \hfill%
    \includegraphics[width=0.33\linewidth]{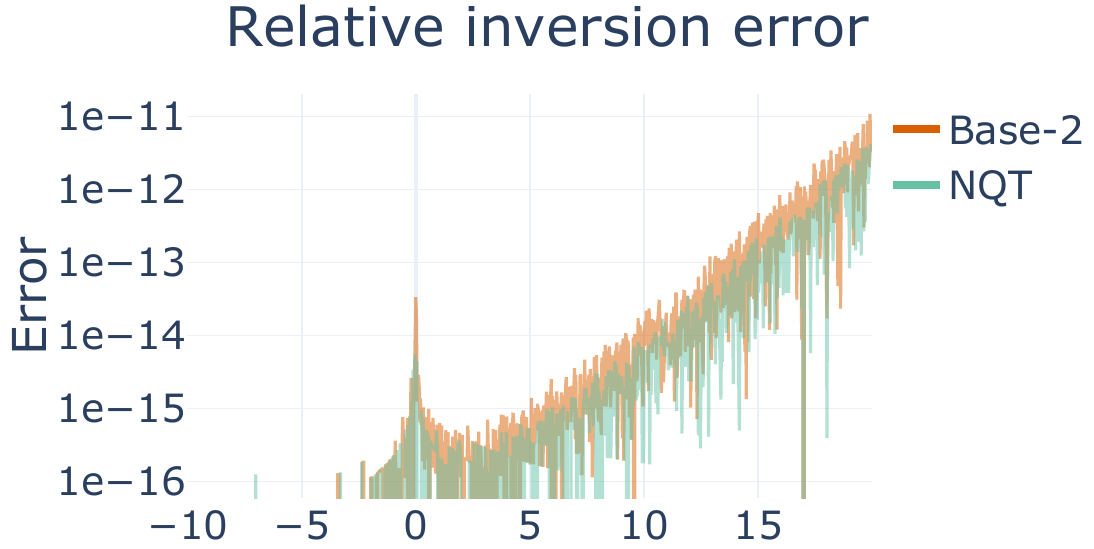}%
    
    \caption{The NQT logistic and logit functions in \zcref{eq:logistic_nqt,eq:logit_nqt} closely follow their standard base-2 counterparts in \zcref{eq:logistic_base2,eq:logit_base2}. The differences are only noticeable when zooming in. Moreover, the NQT logit is the inverse of the NQT logistic with a numerical accuracy on par with the base-2 counterparts.}
    \label{fig:nqt_approximation}
\end{figure}

The NQT nonlinearities share the parameters $\alpha \in \Real_+$ and $x_0 \in [x_{\text{min}}, x_{\text{max}}]$ with the logistic/logit nonlinearities. We therefore use the projections in \zcref{eq:project_logistic} in \zcref{line:nes_projection} of \zcref{algo:SNES} and the same initial values $\vect{\mu}_{\text{init}} = [10, 0]$ and $\sigma_{\text{init}} = [2, 0.5]$.

\paragraph{Efficiency:}
The NQT nonlinearities, in contrast to the logistic/logit functions, do not rely on fundamental functions (exponentials and logarithms) for their computation and can be implemented with a handful of elementary operations, see \zcref{tab:computational_costs}. Implementation details are provided in \zcref{sec:nqt_code}.

\section{Evaluation}
\label{sec:results}

We now evaluate the ability of NVQ to provide computationally-efficient low-loss quantization.  First, we examine how effectively NVQ can reduce loss versus the uniform quantization baseline.  Next, we assess the utility of using per-vector quantizers.  We then examine the effect that both non-linearity selection and subvector count have on loss.  Finally, we evaluate metrics that are important in production deployment, including query recall versus latency and storage footprint savings.

Throughout the experiments, we use the shorthand Log-Log and NQT to refer to the logistic/logit and NQT logistic/logit nonlinearities, respectively. Before quantizing, we center the data using the mean vector of the dataset $\set{X}$, i.e., we subtract the vector $\bar{\vect{x}} = \sum_{\vect{x} \in \set{X}} \vect{x}$. This type of centering has proven effective and robust to distribution shifts~\citep{aguerrebere_similarity_2023}. We use the datasets described in \zcref{tab:datasets}, which stem from different embedding models and types of data.

\begin{table}[t]
    \caption{Datasets used for our experimental results. We denote the dimensionality and the number of vectors in the dataset by $d$ and $n$, respectively.}
    \label{tab:datasets}

    %

    \begin{minipage}{0.32\linewidth}
        \centering
        \small
        \begin{tabular}{lcc}
            \toprule
            Name & $d$ & $n$ \\
            \midrule
            ada-002-100k\footnote{\label{jvector-url}\url{https://github.com/datastax/jvector}} &  1536 & $10^5$ \\
            openai-v3-100k\footref{jvector-url} & 1536 & $10^5$ \\
            gecko-100k\footref{jvector-url} & 768 & $10^5$ \\
            \bottomrule
        \end{tabular}        
    \end{minipage}%
    \hfill
    \begin{minipage}{0.32\linewidth}
        \centering
        \small
        \begin{tabular}{lcc}
            \toprule
            Name & $d$ & $n$ \\
            \midrule
            dpr-1M~\citep{aguerrebere_similarity_2023} & 768 & $10^6$ \\
            cohere-1M\footnote{\label{cohere-url}\url{https://huggingface.co/datasets/Cohere/wikipedia-22-12}} & 1024 & $10^6$ \\
            cap-1M\footnote{\label{caselaw-url}\url{https://huggingface.co/datasets/laion/Caselaw_Access_Project_embeddings}} & 1536 & $10^6$ \\
            \bottomrule
        \end{tabular}
    \end{minipage}%
    \hfill
    \begin{minipage}{0.32\linewidth}
        \centering
        \small
        \begin{tabular}{lcc}
            \toprule
            Name & $d$ & $n$ \\
            \midrule
            dpr-10M~\citep{aguerrebere_similarity_2023} & 768 & $10^7$ \\
            cohere-10M\footref{cohere-url} & 1024 & $10^7$ \\
            cap-6M\footref{caselaw-url} & 1536 & $6.07 \cdot 10^6$ \\
            \bottomrule
        \end{tabular}
    \end{minipage}
\end{table}

\subsection{Loss reduction and convergence}
\label{sec:loss_reduction}

\begin{figure}[t]
    \centering
    \hfill%
    \includegraphics[width=0.33\linewidth]{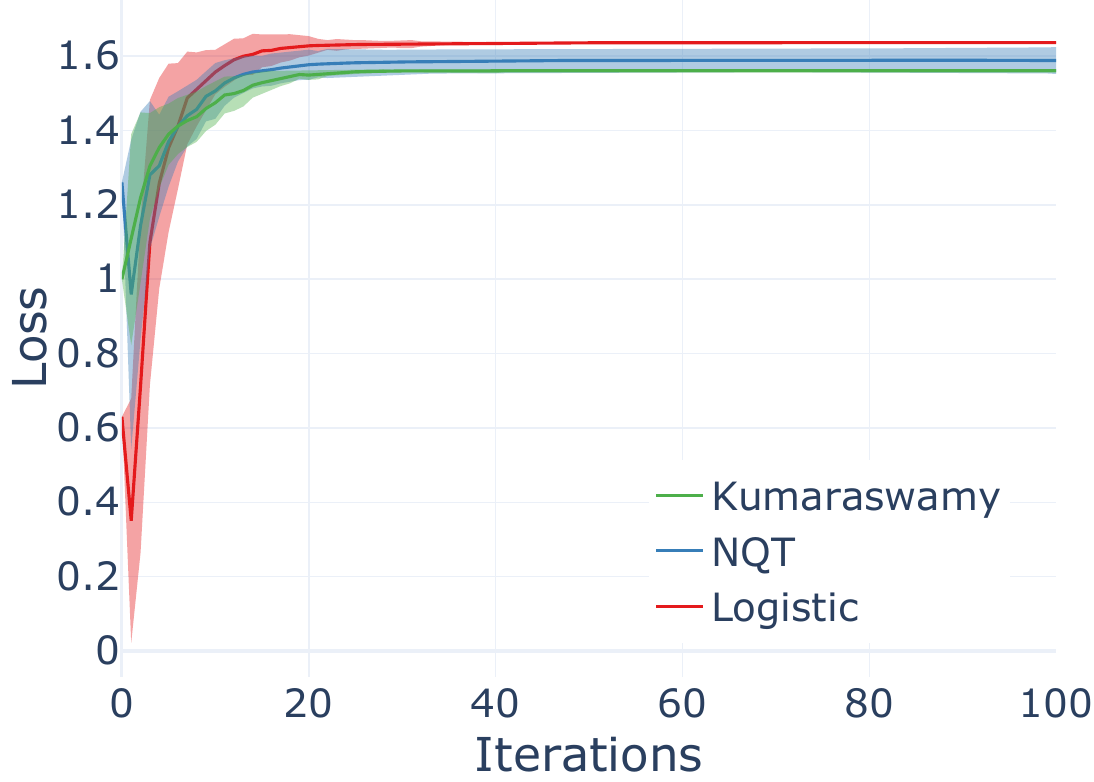}%
    \hfill%
    \includegraphics[width=0.33\linewidth]{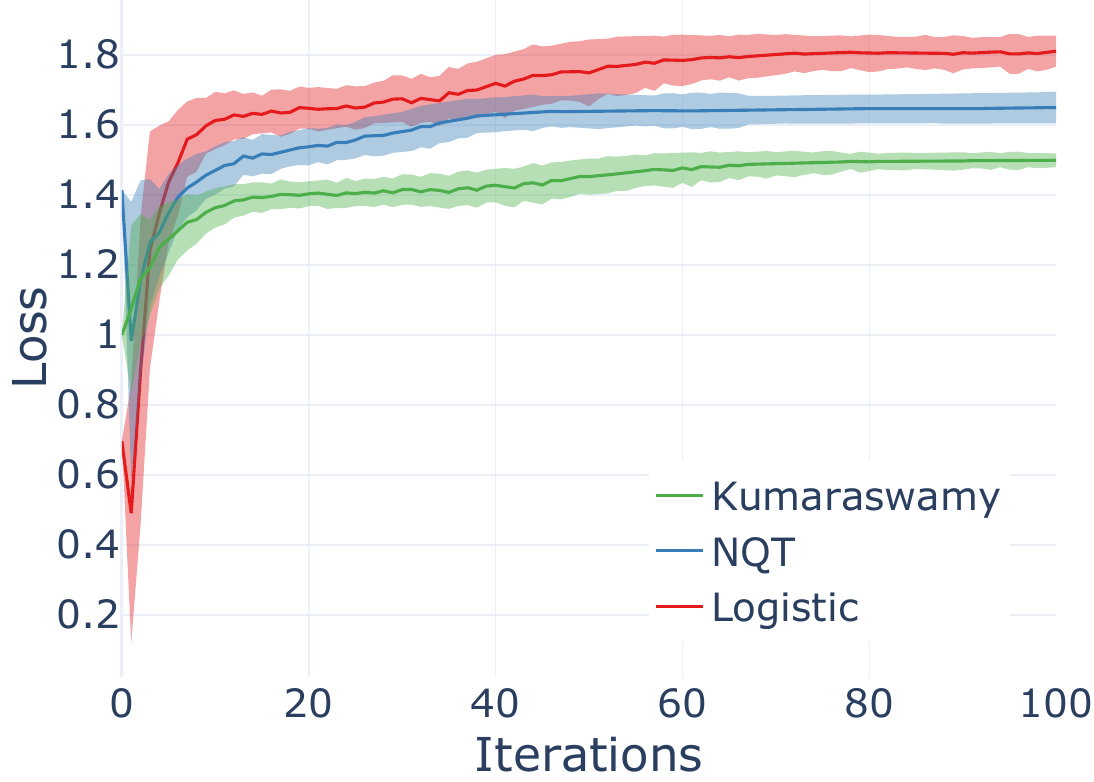}%
    \hfill%
    \includegraphics[width=0.33\linewidth]{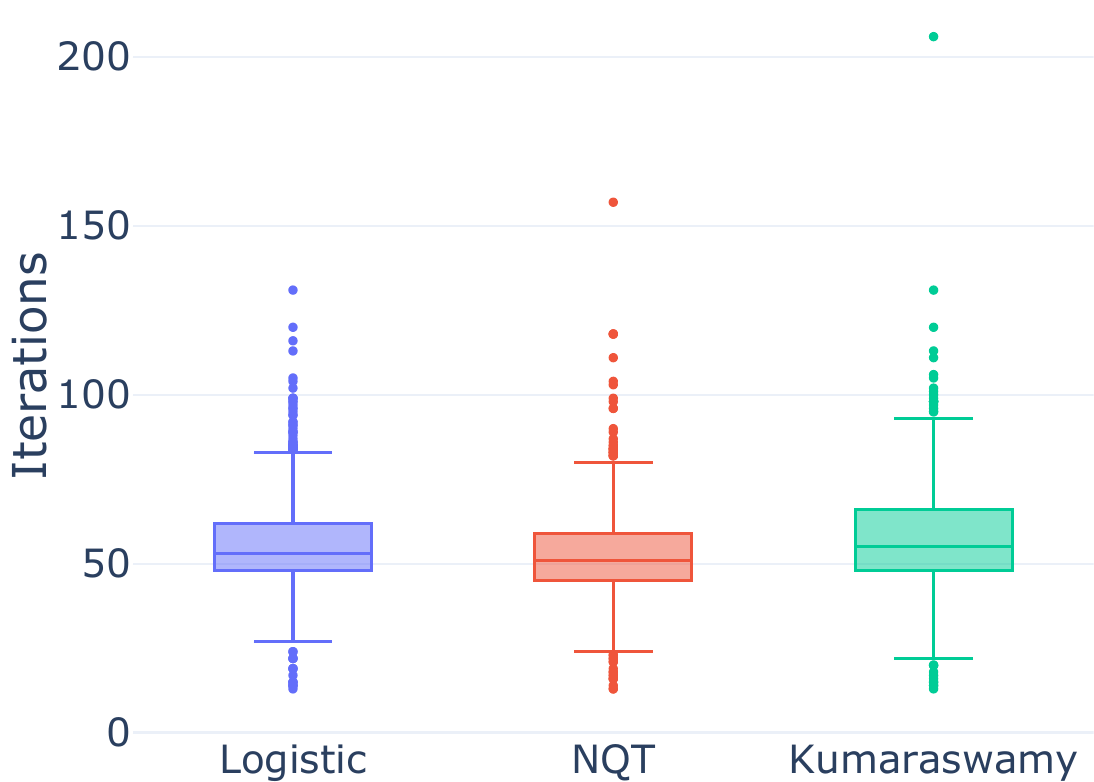}%
    \caption{\textbf{Left and center:} Evolution of the objective function (higher is better) in Problem~\zcref[noname]{prob:nuveq} for different runs of \zcref{algo:SNES} (the shaded region indicates plus/minus one standard deviation due to the stochasticity of the natural gradient) for 4-bit (left) and 8-bit (right) NVQ.
    \textbf{Right:} The number of iterations of \zcref{algo:SNES} to reach the stopping condition in \zcref{sec:optimization} (i.e., an absolute error of $10^{-4}$ across consecutive iterations) with $\beta=8$ bits for vectors in ada002-100k. On average, 50 iterations suffice.}
    \label{fig:convergence}
\end{figure}

We start by examining the capability of \zcref{algo:SNES} to optimize Problem~\zcref[noname]{prob:nuveq}.
In the left and center plots of \zcref{fig:convergence}, we analyze the convergence of several runs of \zcref{algo:SNES} for a single vector. 
For this, we remove the stopping condition in \zcref{sec:optimization} and run the algorithm for a fixed number of 100 iterations. Since SNES uses a stochastic estimate of the natural gradient, we observe relatively small variability across different runs of the algorithm. All runs are successful and achieve an improvement over the baseline (recall that the objective function in Problem~\zcref[noname]{prob:nuveq} is the relative improvement over the uniform baseline). Few iterations are required to reach convergence. After 20 iterations, the algorithm converges for 4 bits and minor improvements are achieved for 8 bits (no noticeable change after 50 iterations). This is corroborated in right plot, where we observe that 50 iterations are sufficient on average (computed across vectors) to reach an absolute error of $10^{-4}$ between two consecutive iterations.

\subsection{Value of per-vector quantizers}
\label{sec:per_vector_quantizers}

\begin{figure}[t]
    \centering
    \includegraphics[width=0.32\linewidth]{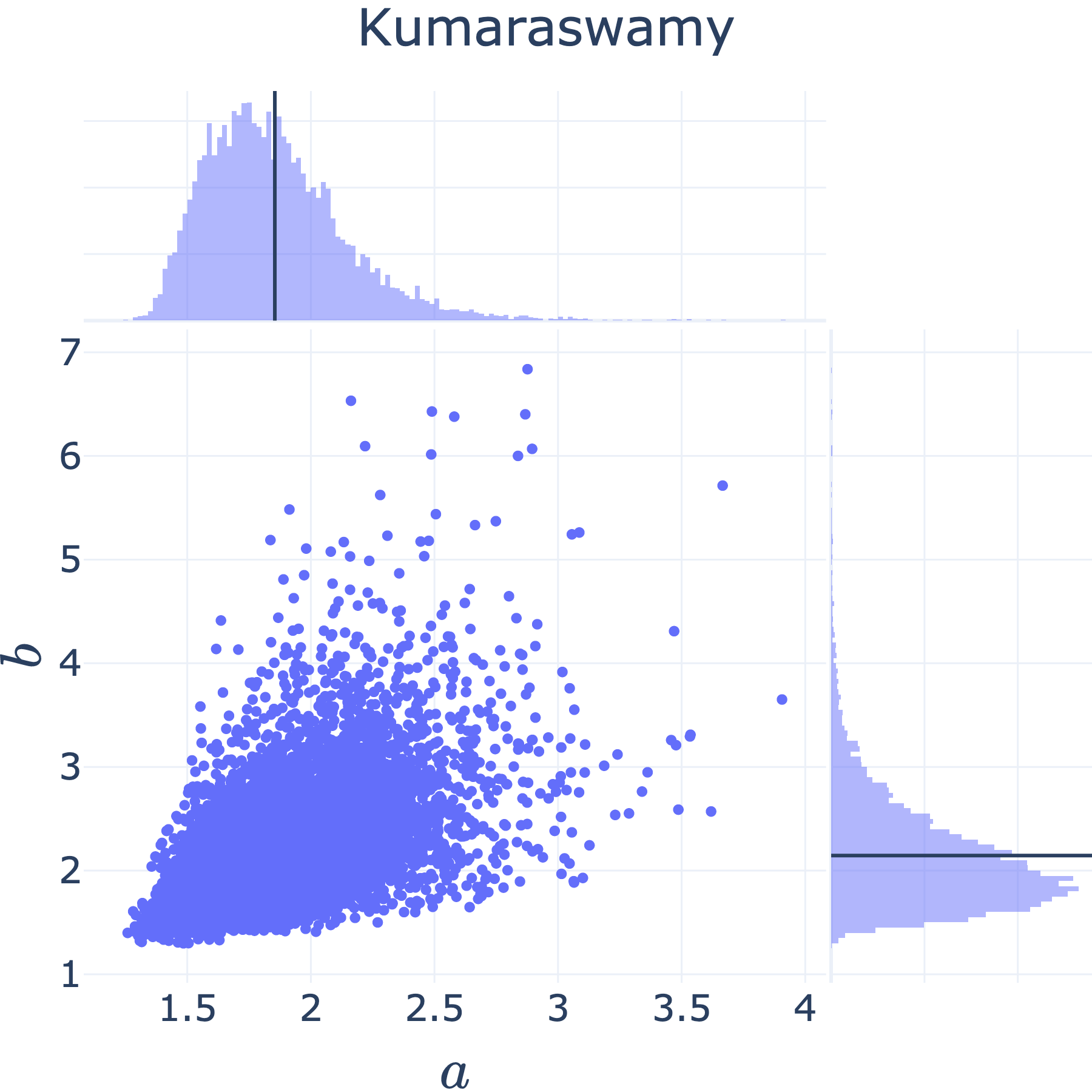}%
    \hfill%
    \includegraphics[width=0.32\linewidth]{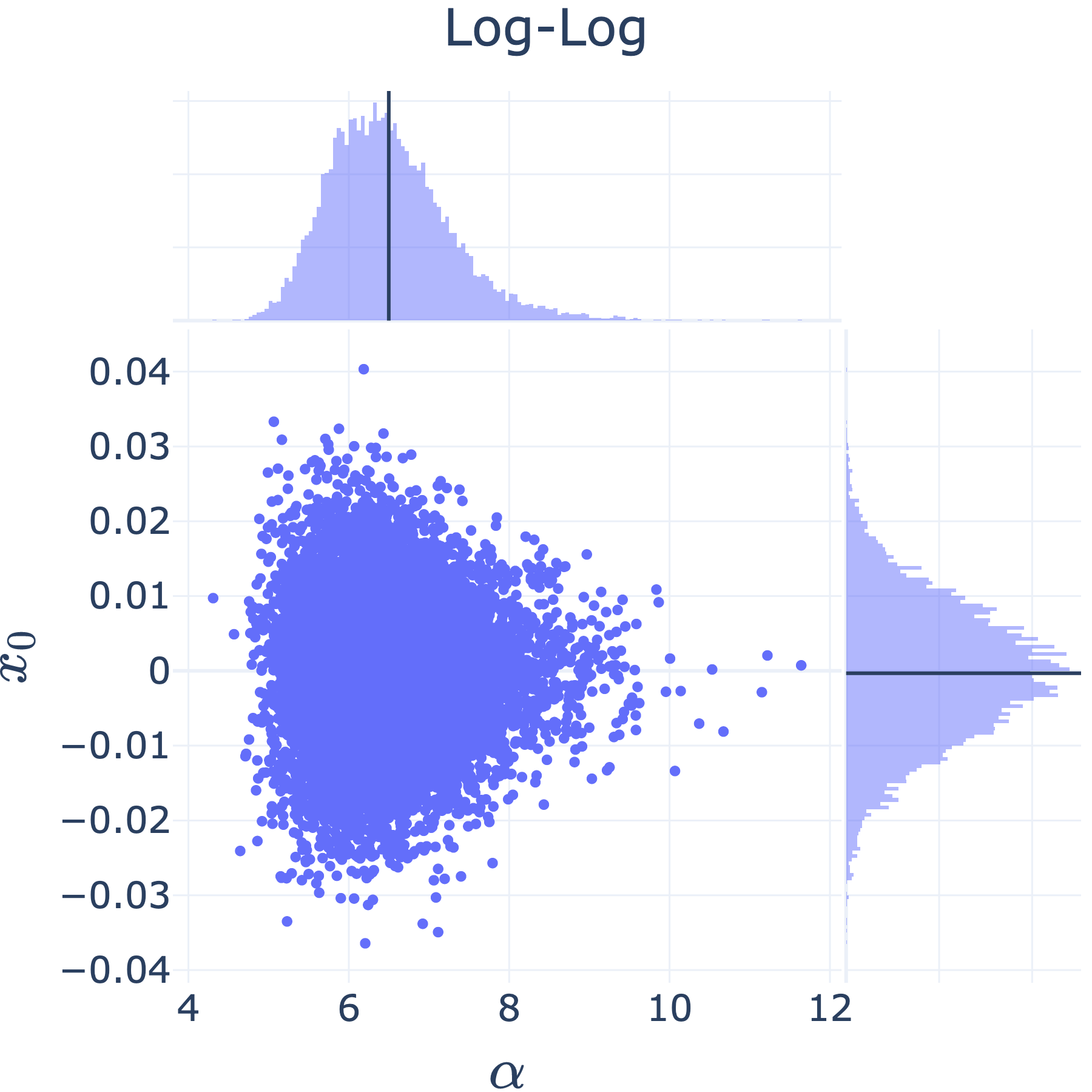}%
    \hfill%
    \includegraphics[width=0.32\linewidth]{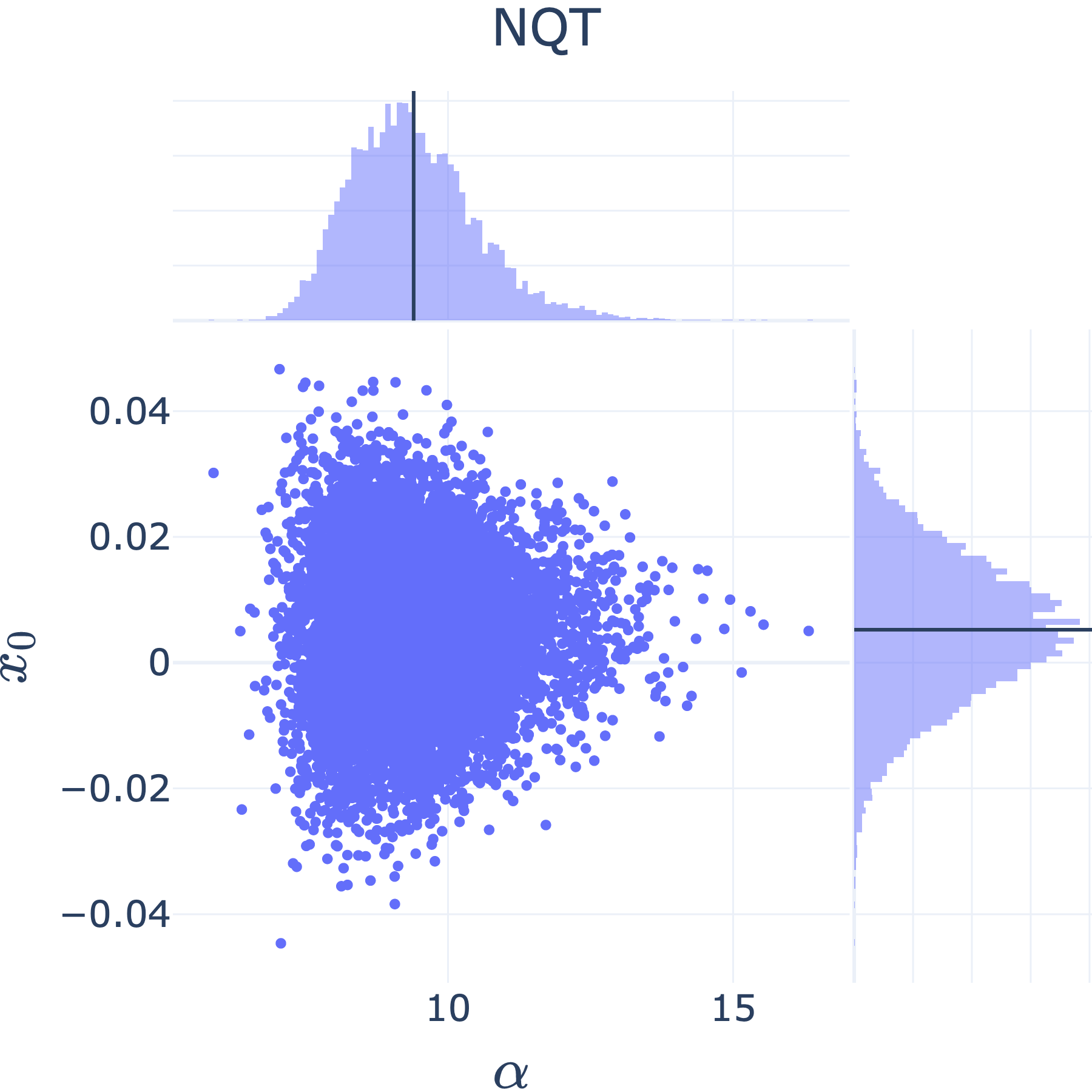}%
    
    \caption{The 8-bit NVQ solutions for $10^4$ vectors (each circle represents the solution for one vector) from ada002-100k. Different nonlinearity parameters are chosen for each vector, justifying the individualized per-vector learning. This pattern is highly repeatable over datasets, as obhserved in \zcref{fig:solutions_distribution_gecko} of the appendix.}
    \label{fig:solutions_distribution_ada002}
\end{figure}

In \zcref{fig:solutions_distribution_ada002} we depict the solutions found by \zcref{algo:SNES} when optimizing Problem~\zcref[noname]{prob:nuveq} for many vectors. Different parameters are chosen for different vectors. This sizable spread in the selected parameters gives grounds for the per-vector approach in NVQ. Additionally, the Log-Log and NQT solutions, being similar in nature, exhibit similar parameter distributions (shifted by the change from the natural base to base 2).

In \zcref{fig:multi_vector} (left), we can observe that each solution found by \zcref{algo:SNES} for different vectors (see \zcref{fig:solutions_distribution_ada002}) is better than what can be achieved with uniform quantization (i.e., no solution is below 1). The Kumaraswamy, Log-Log, and NQT nonlinearities produce improvements of  \mytexttilde$1.81$x, \mytexttilde$1.90$x, and \mytexttilde$1.72$x on average, respectively, while reaching \mytexttilde$4$x for some vectors.

\subsection{Comparison of non-linearities}
\label{sec:non-linearity_comparison}

\begin{figure}[t]
    \centering
    \hfill%
    \includegraphics[width=0.33\linewidth]{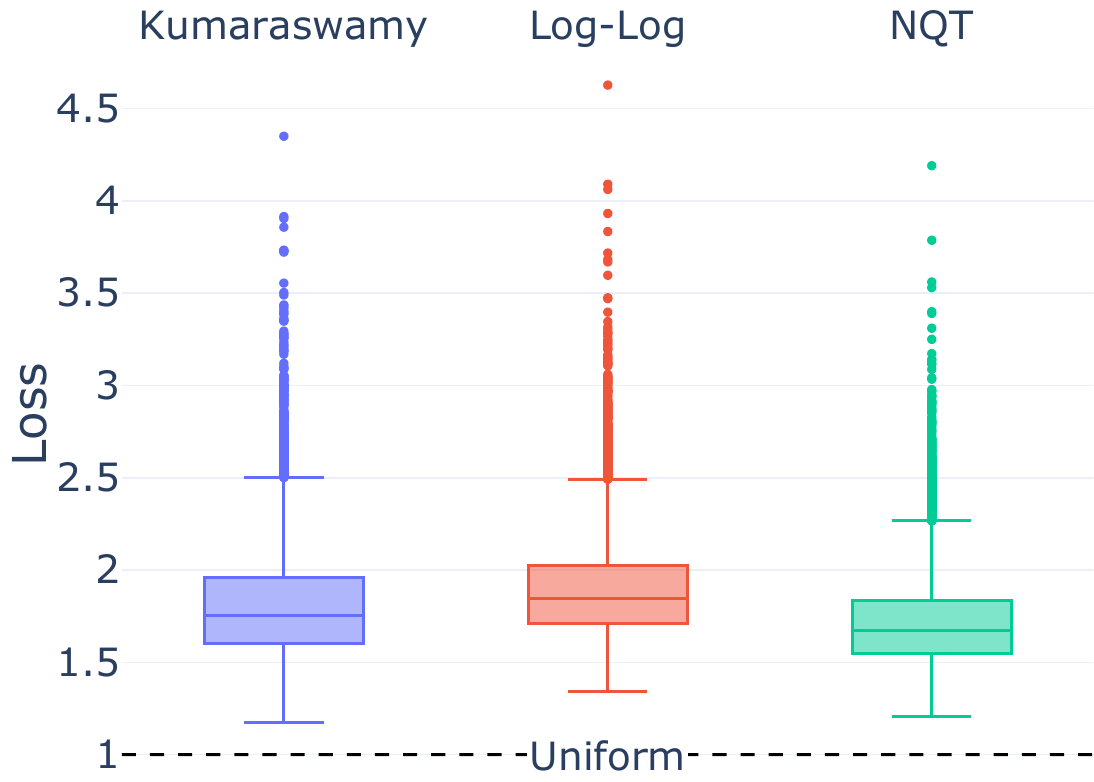}%
    \hfill%
    \includegraphics[width=0.33\linewidth]{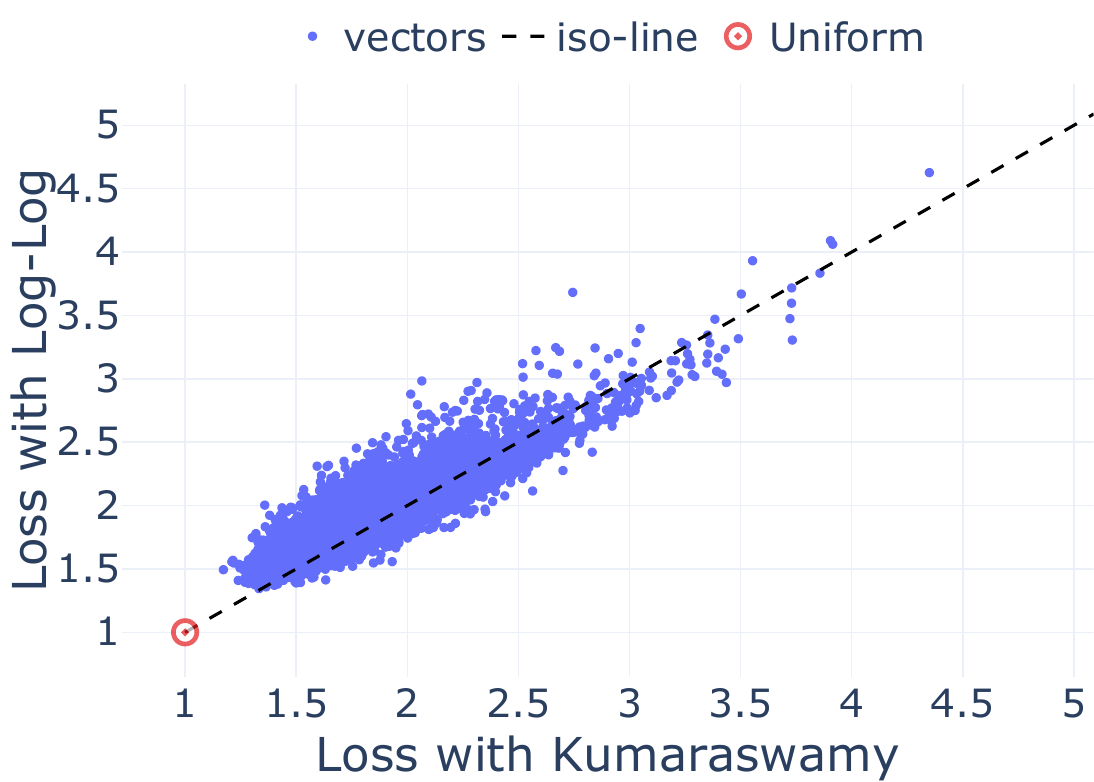}%
    \hfill%
    \includegraphics[width=0.33\linewidth]{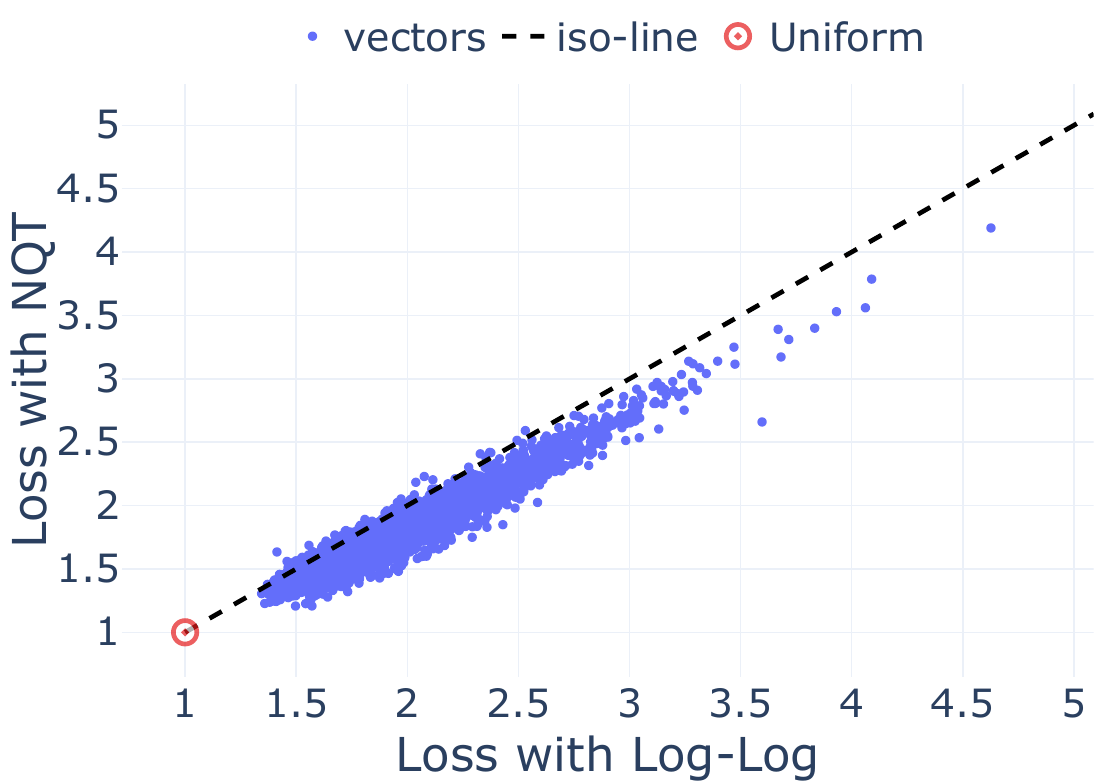}%
    
    \caption{\textbf{(Left)} The objective function values achieved when maximizing Problem~\zcref[noname]{prob:nuveq} with $\beta=8$ bits for $10^4$ vectors from ada002-100k (the objective function is the relative improvement over the uniform scalar quantization). On average, we achieve an improvement of \mytexttilde$1.8$x (higher is better). There are no vectors that see no improvement.
    \textbf{(Center and right)} Comparison of the objective function values achieved when optimizing Problem~\zcref[noname]{prob:nuveq} with $\beta=8$ bits using different nonlinearities for $10^4$ vectors from ada002-100k (each vector is represented by a circle). At the iso-line, both nonlinearities are equally good. Log-Log is slightly better for most vectors compared to the Kumaraswamy (center) and NQT (right) nonlinearities.}
    \label{fig:multi_vector}
\end{figure}

We also individually compare the objective function values of the solutions found with different nonlinearities in \zcref{fig:multi_vector} (center and right). In general, Log-Log is slightly better than the Kumaraswamy and NQT nonlinearities for vectors stemming for commonly used embedding models. However, as usual in machine learning, there is no silver bullet: the NQT nonlinearity is faster to compute but the Kumaraswamy nonlinearity offers more representational flexibility.

\subsection{Benefit of subvectors}
\label{sec:subvector_benefit}

\begin{figure}[t]
    \centering
    \includegraphics[width=0.9\linewidth]{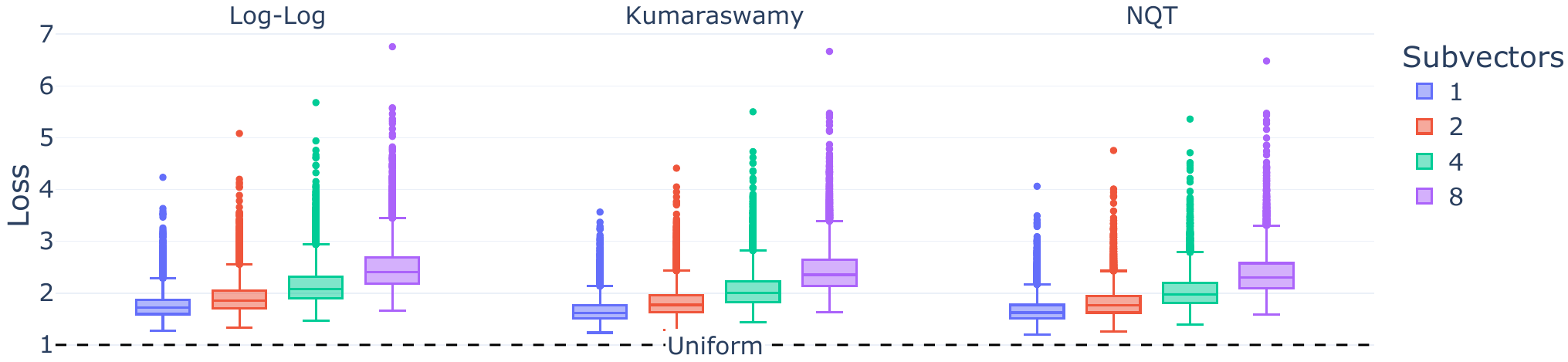}
    
    \includegraphics[width=0.9\linewidth]{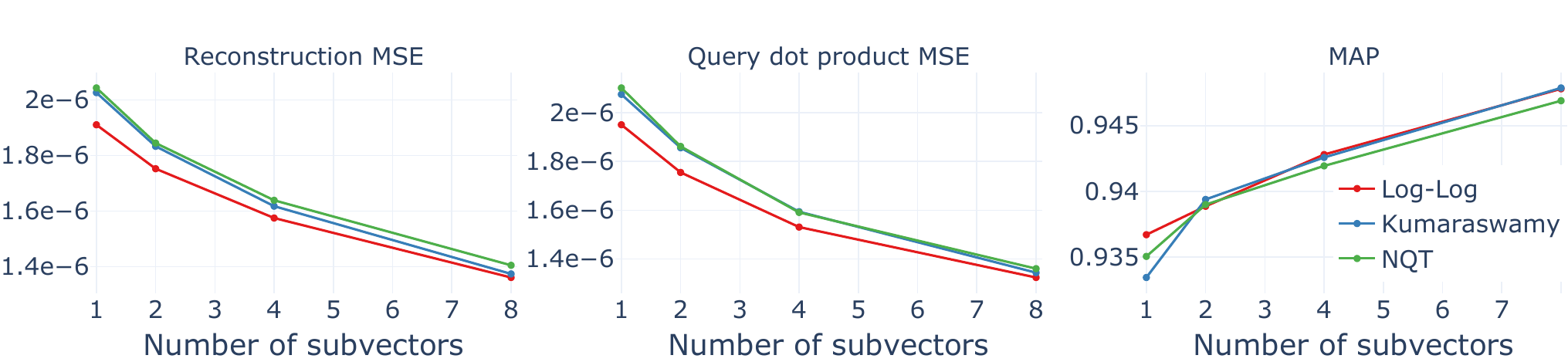}
    
    \caption{(\textbf{Top}) Increasing the number of subvectors increases the objective function (higher is better) in Problem~\zcref[noname]{prob:nuveq}. There are no vectors that see no improvement over the baseline.
    (\textbf{Bottom}) The average reconstruction error and the query dot product error, defined as $\sum_{\vect{x} \in \set{X}}(\langle \vect{q}, \vect{x} \rangle - \langle \vect{q}, \widetilde{\vect{x}} \rangle )^2$ decrease (lower is better) and the mean average precision (MAP) increases (higher is better).
    Results obtained with $\beta=4$ for $10^4$ vectors from ada002-100k.
    Other datasets and values of $\beta$ are in \zcref{fig:subvectors_loss_continued,fig:subvectors_recall_continued} of the appendix.
    }
    \label{fig:subvectors}
\end{figure}

We study the effects of changing the number of subvectors in \zcref{fig:subvectors}. We observe that using more subvectors improves the value of the objective function in Problem~\zcref[noname]{prob:nuveq}. This is natural as the domain $[x_{\text{min}}, x_{\text{max}}]$ corresponmdig to each subvector becomes tighter as the number of subvectors increases. This effect is clearly reflected in other vector search metrics, as depicted in the bottom row of \zcref{fig:subvectors}. These improvements come with a slightly increase of the payload of each compressed vector, as described in \zcref{sec:subvectors}.

\subsection{Evaluation of search storage footprint}
\label{sec:storage_footprint}

\begin{figure}[t]
    \centering
    \includegraphics[width=0.33\linewidth]{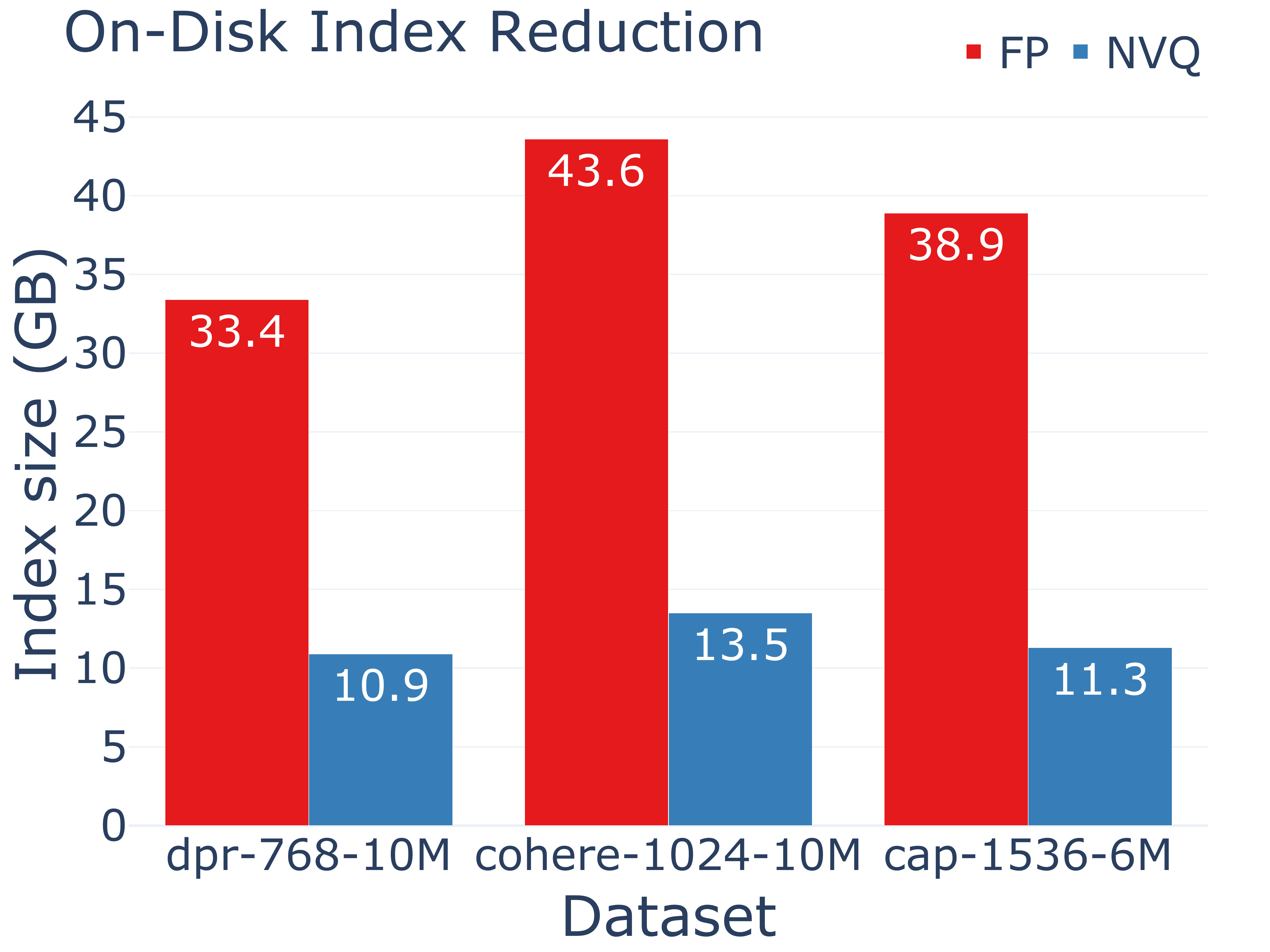}
    \caption{The on-disk vector index size for FP32 vectors (red) and NVQ vectors (blue), for larger datasets of 768, 1024, and 1536 dimensions. In thse cases, NVQ achieves storage savings of 3.06x, 3.23x, and 3.44x with a minute loss in accuracy (see \zcref{fig:recall_vs_latency}).}
    \label{fig:index_footprint_hist}
\end{figure}

As described in \zcref{sec:related}, vector quantization is commonly used to conserve memory and accelerate search at the expense of recall.  Reranking search results with full-precision vectors can mitigate the impact but requires random reads from storage of a small multiple of $k$ vectors for top-$k$ retrieval.  For a production system servicing thousands of queries/sec, the storage costs are driven by the amount of high-performance capacity required, which can be expensive for large-scale datasets, making this the 3rd largest system cost after compute and memory. 

We measure the impact that NVQ has on the index storage capacity requirements in \zcref{fig:index_footprint_hist}.  Shown is the storage footprint in terms of measured file size for datasets encoded using 8-bit NVQ with two subvectors versus the original FP32 encoding.  NVQ uses 3.06x, 3.23x, and 3.44x less storage than FP32 for the 768-, 1024-, and 1536-dimensional datasets, respectively. The savings are higher for higher dimensionality due to the reduced format overhead.

\subsection{Evaluation of search speed}
\label{sec:search_speed}

\begin{figure}[t]
    \centering
    \hfill%
    \includegraphics[width=0.33\linewidth]{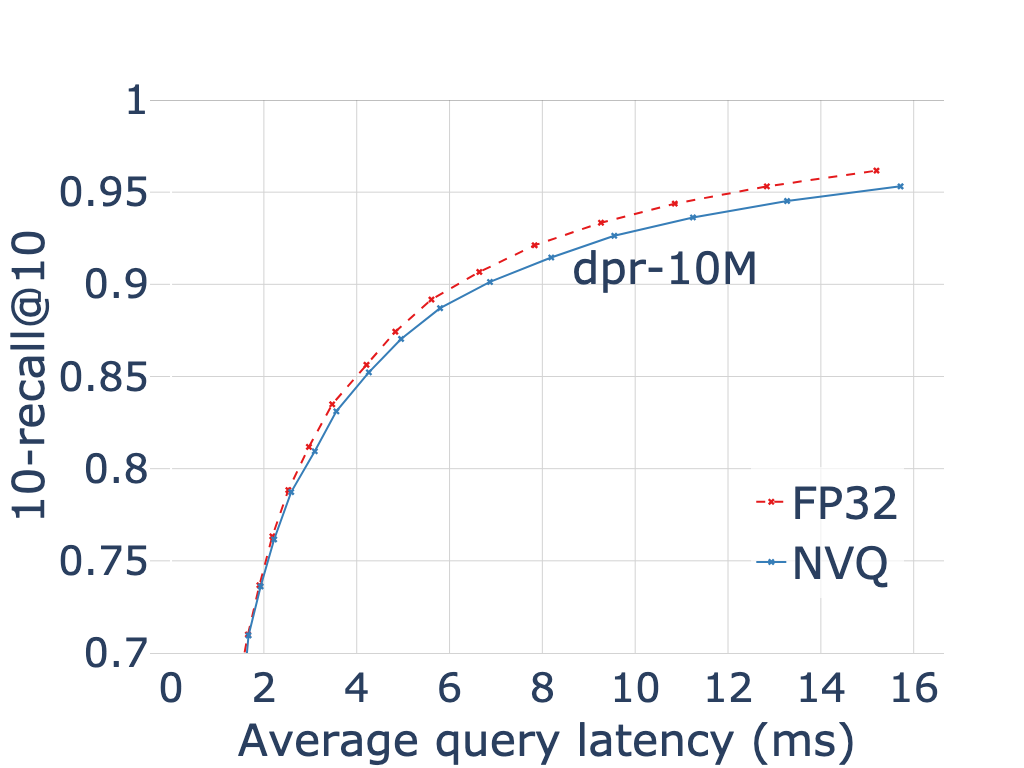}%
    \hfill%
    \includegraphics[width=0.33\linewidth]{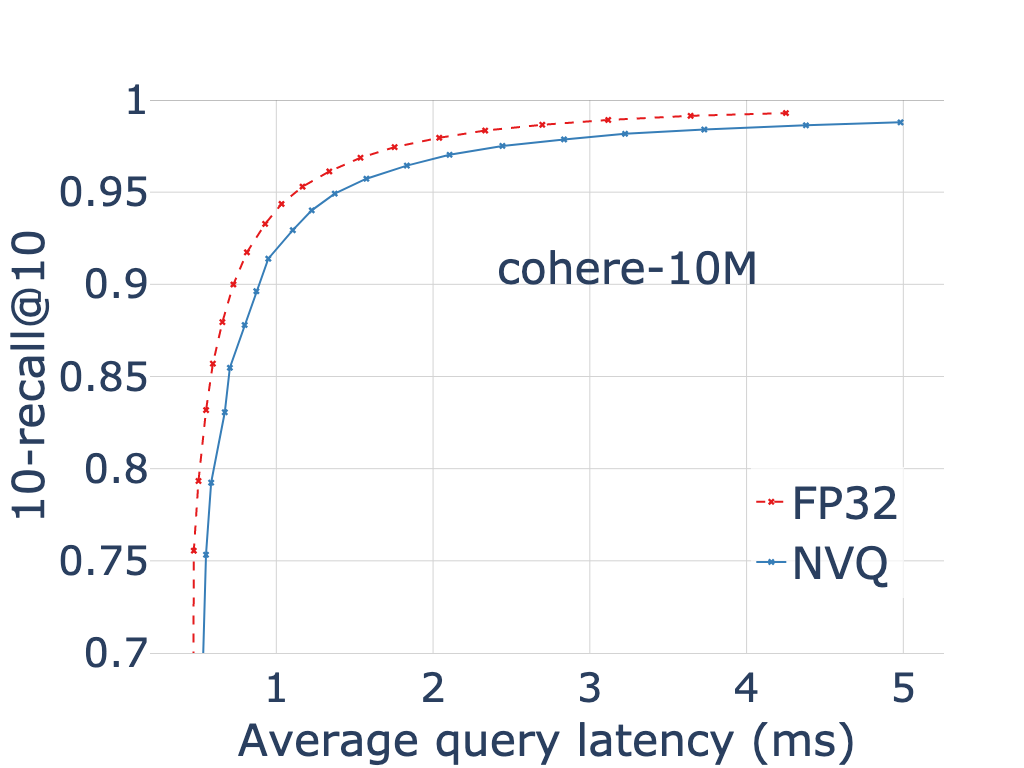}%
    \hfill%
    \includegraphics[width=0.33\linewidth]{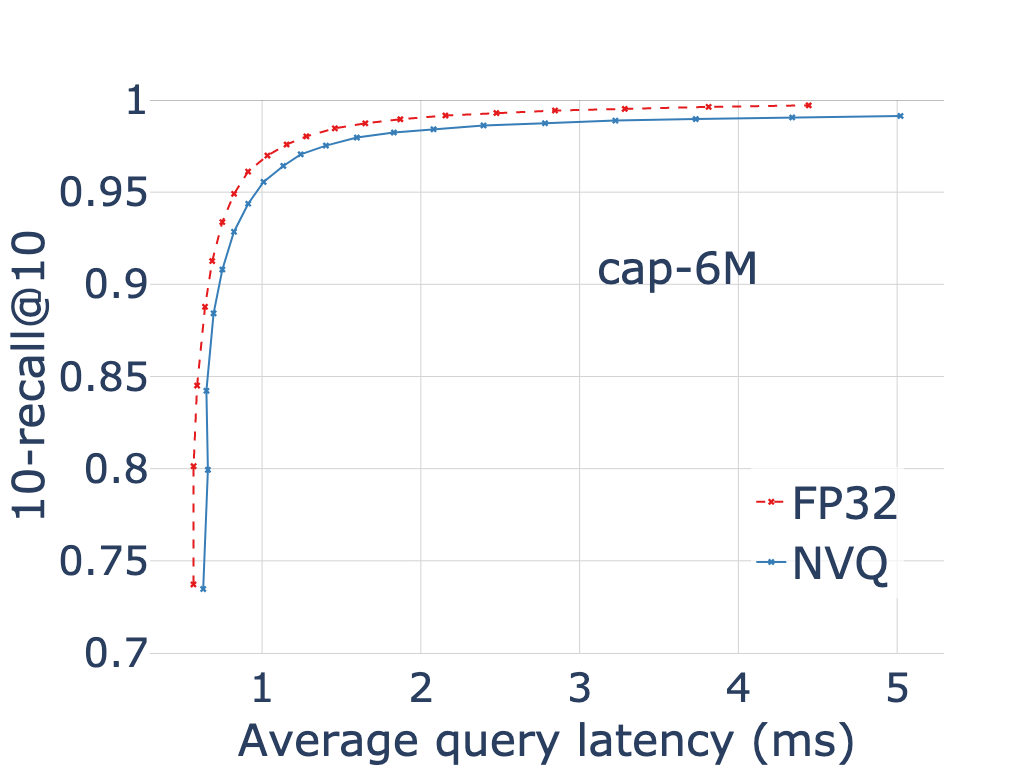}%
    
    \caption{Recall versus query latency for three larger datasets.  A graph-based index was constructed using PQ vectors.  Reranking was then performed using either NVQ or FP32 vectors.  The results show that NVQ has a very small deleterious effect on recall in the high-recall regime for 768 dimensions \textbf{(Left)}, 
     1024 dimensions \textbf{(Center)},
     and 1536 dimensions \textbf{(Right)}.
    }
    \label{fig:recall_vs_latency}
\end{figure}

Finally, we look at search recall versus latency on several larger datasets with a range of dimensionality from 768 to 1536, a popular range for today's retrieval-augmented generation systems.  This evaluation exposes the impact of NVQ reconstruction loss in terms of degraded recall as well as the cost of decoding the quantized vectors in terms of latency.  A graph index was constructed using the Vamana algorithm described in~\citep{subramanya_diskann_2019} (out-degree $R=64$, search window size of 200, $\alpha=1.2$) and vectors encoded using Product Quantization (256 centroids, 16x compression).  Queries were then performed from a held-out query set of 10k queries while the amount of reranking was swept from 1.0x (i.e., top-k PQ vectors reranked with k FP or NVQ vectors) to a sufficiently-high ratio (e.g., 40-80x).  In the high-recall regime of most interest ($>0.95$), 10-recall@10 is impacted less than $0.01$ at the same latency.

\section{Conclusion}
\label{sec:conclusions}

We introduced Non-uniform Vector Quantization (NVQ) for high-fidelity compression of vectors for similarity search at modern embedding dimensionalities.  The key idea is to replace a one-size-fits-all uniform scalar quantizer with an \emph{individualized}, two-parameter, non-uniform quantizer per vector (or subvector), chosen by a tiny optimization problem at insert time.  We showed that three parsimonious nonlinearities- Kumaraswamy, scaled logistic/logit, and non-quite-transcendental logistic/logit- span a useful representational-speed Pareto front, and that a lightweight, gradient-free optimizer reliably finds good settings in a few dozen iterations.  Empirically in both 4- and 8-bit regimes, NVQ consistently reduces reconstruction loss relative to uniform scalar quantization by 1.7 to 1.9x, reaching ~4x for some vectors.  Subvectorization into 2,4, or 8 subvectors further improves reconstruction with minimal overhead.  When NVQ is used to quantize vectors for reranking, NVQ can reduce storage use by 3.06 to 3.44x versus 32-bit full-precision vectors for 768 to 1536 dimensional data, respectively, with an impact of less than 0.01 on recall above 0.95.

NVQ also has limitations that point to promising directions. First, we optimized mean‑squared reconstruction error as a surrogate for ranking quality; incorporating query‑aware or rank‑based losses could further tighten recall–latency trade‑offs (e.g., see~\citep{tepper_leanvec_2024}). Second, our normalization relies on per‑(sub)vector min/max and may be sensitive to outliers; robust or learned normalizers are worth exploring. Third, while we evaluated 4-8 bit data types, pushing into ultra‑low bitrates will likely require additional structure (e.g., mixed precision across dimensions) or multi‑level schemes. Finally, avoiding full dequantization during reranking by deriving unbiased dot‑product estimators in the quantized domain (in the spirit of the asymmetric distance computation from~\citep{jegou_product_2011}) could reduce compute and memory traffic further, especially on accelerators.

\clearpage

\printbibliography

\appendix

\clearpage

\section{Fast logarithms and exponentiation}
\label{sec:fast_exponentiation}

Even if the expressions of $F_{\textsc{ks}}$ and $F_{\textsc{ks}}^{-1}$ are mathematically straightforward, the use of fundamental functions makes their computation challenging.
To speedup these computations, we use the identity
\begin{equation}
    x^c = \exp{ \left( c \log{x} \right) } ,
\end{equation}
enabling the use of fast approximations of the common functions $\exp$ and $\log$ for $x \in \Real_+$.

For the exponential and logarithm, we use an implementation based on the minimax polynomial approximation.\footnote{\url{https://stackoverflow.com/a/39822314} and \url{https://stackoverflow.com/a/47025627}} We have modified the Remez polynomials for an optimal tradeoff between speed and accuracy using \url{https://github.com/DKenefake/OptimalPoly/}.
We reproduce the code for completeness.

\begin{minted}{c++}
#include <cmath>
#include <cstdint>
#include <cstring>

float int_as_float (int32_t a) {
    float r;
    memcpy (&r, &a, sizeof r);
    return r;
}
int32_t float_as_int (float a) {
    int32_t r;
    memcpy (&r, &a, sizeof r);
    return r;
}

float fast_logf (float a) {
    float m, r, s, t, i, f;
    int32_t e;

    e = (float_as_int(a) - 0x3f2aaaab) & 0xff800000;
    m = int_as_float(float_as_int(a) - e);
    i = (float) e * 1.19209290e-7f;
    f = m - 1.0f;
    s = f * f;
    r = fmaf(0.230836749f, f, -0.279208571f);
    t = fmaf(0.331826031f, f, -0.498910338f);
    r = fmaf(r, s, t);
    r = fmaf(r, s, f);
    r = fmaf(i, 0.693147182f, r);
    return r;
}

float fast_exp(float x) {
    float invlog2e = 1.442695041f;  // 1 / log2(e)
    float expCvt = 12582912.0f;  // 1.5 * (1 << 23)

    /* exp(x) = 2^i * 2^f; i = rint (log2(e) * x), -0.5 <= f <= 0.5 */
    float t = x * invlog2e;  // t = x / log2(e)
    float r = (t + expCvt) - expCvt;  // r = round(t)
    float f = t - r;  // f = t - round(t)
    int i = (int) r; // i = (int) r

    float temp = fmaf(f, 0.009651907610706037f, 0.05593479631997887f);
    temp = fmaf(temp, f, 0.2402301551437674f);
    temp = fmaf(temp, f, 0.6931186232012877f);
    temp = fmaf(temp, f, 0.9999993887682104f);

    temp = int_as_float(float_as_int(temp) + (i << 23));  // temp = temp * 2^i
    return temp;
}
\end{minted}

\section{Fast NQT logistic and logit functions}
\label{sec:nqt_code}

\begin{minted}{c++}
#include <cmath>
#include <cstdint>
#include <cstring>

float int_as_float (int32_t a) {
    float r;
    memcpy (&r, &a, sizeof r);
    return r;
}
int32_t float_as_int (float a) {
    int32_t r;
    memcpy (&r, &a, sizeof r);
    return r;
}

float logisticNQT(float value, float alpha, float x0) {
    // The value -alpha * x0 can be precomputed
    float temp = fmaf(value, alpha, -alpha * x0);
    int p = round(temp + 0.5f);
    int m = float_as_int(fmaf(temp - p, 0.5f, 1f));
    
    temp = int_as_float(m + (p << 23));  // temp = m * 2^p
    return temp / (temp + 1f);
}

float logitNQT(float value, float inverseAlpha, float x0) {
    float z = value / (1f - value);
    
    int temp = float_as_int(z);
    int e = temp & 0x7f800000;
    float p = (float) ((e >> 23) - 128);
    float m = int_as_float((temp & 0x007fffff) + 0x3f800000);
    
    return fmaf(m + p, inverseAlpha, x0);
}
\end{minted}

\section{Additional experimental results}

\begin{figure}[p]
    \centering
    \begin{tblr}{
        colspec = {X[c,r,1]X[c,h,10]X[c,h,10]},
    }
        \begin{sideways}
            \small Kumaraswamy
        \end{sideways} &
        \includegraphics[width=\linewidth,align=c]{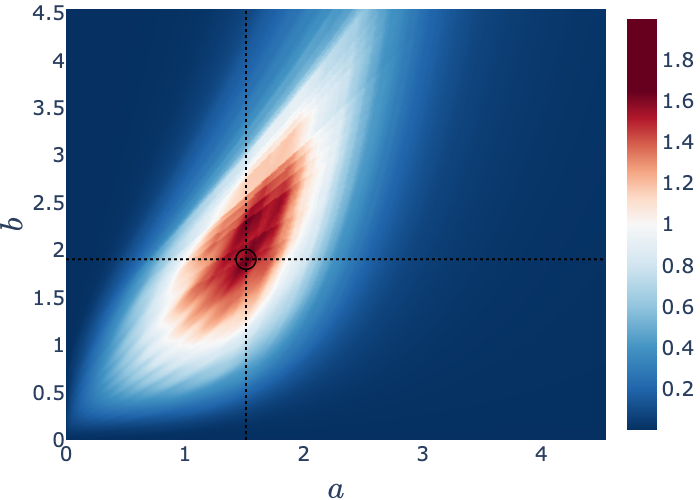} &
        \includegraphics[width=\linewidth,align=c]{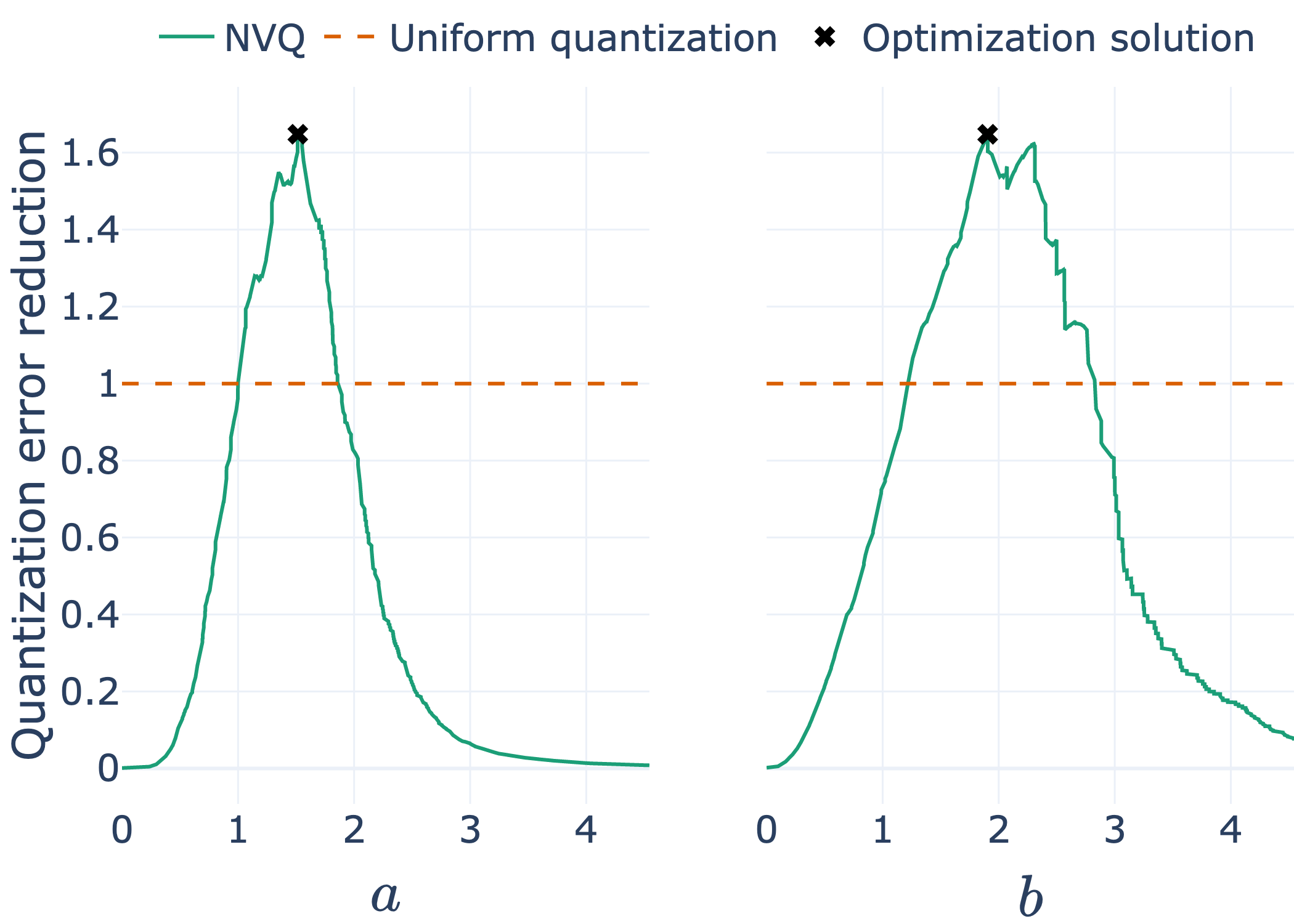} \\

        \begin{sideways}
            \small Log-Log
        \end{sideways} &
        \includegraphics[width=\linewidth,align=c]{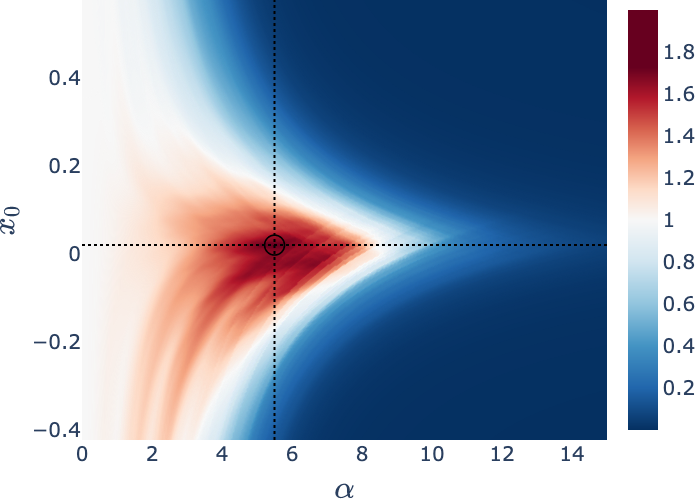} &
        \includegraphics[width=\linewidth,align=c]{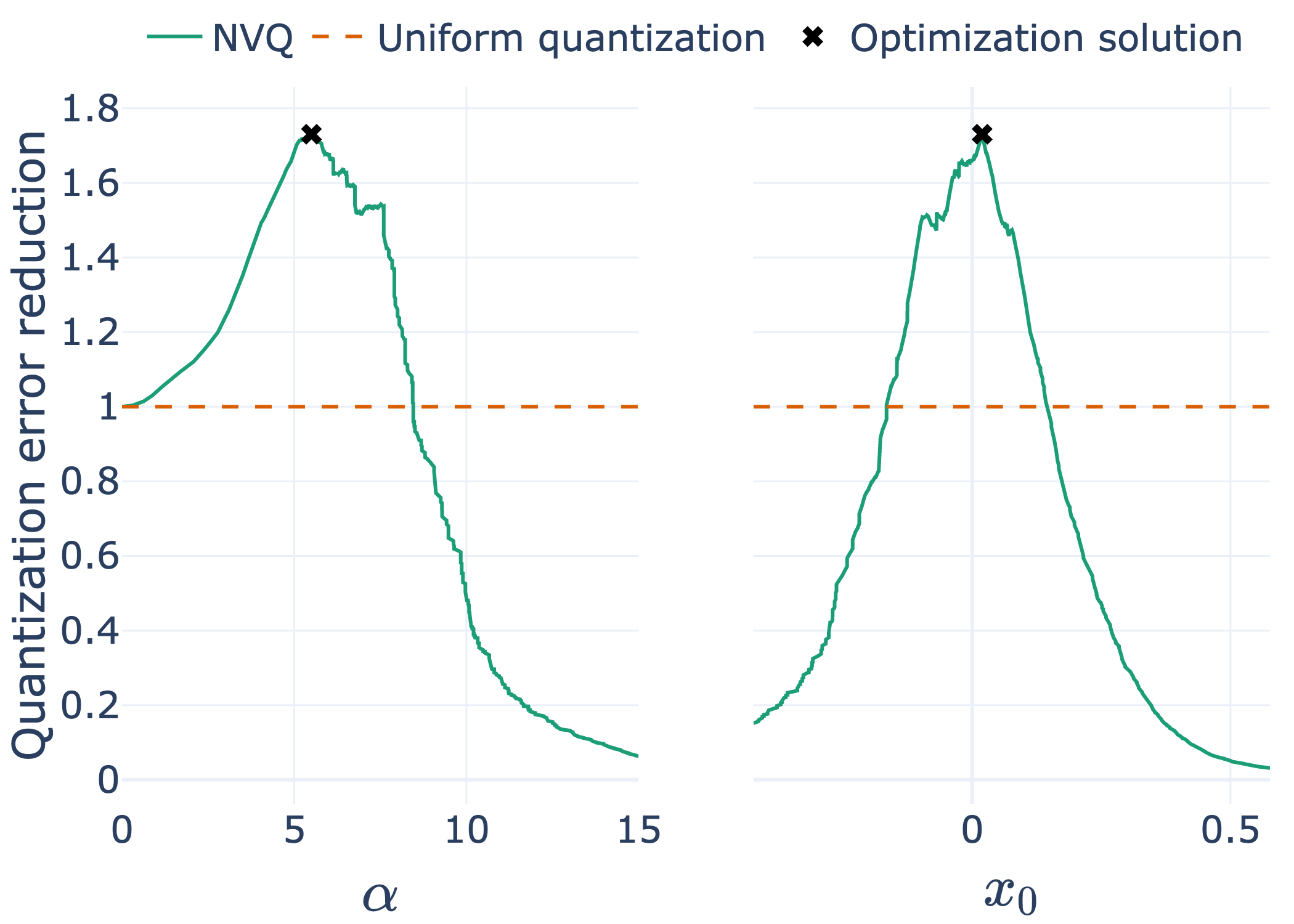} \\

        \begin{sideways}
            \small NQT
        \end{sideways} &
        \includegraphics[width=\linewidth,align=c]{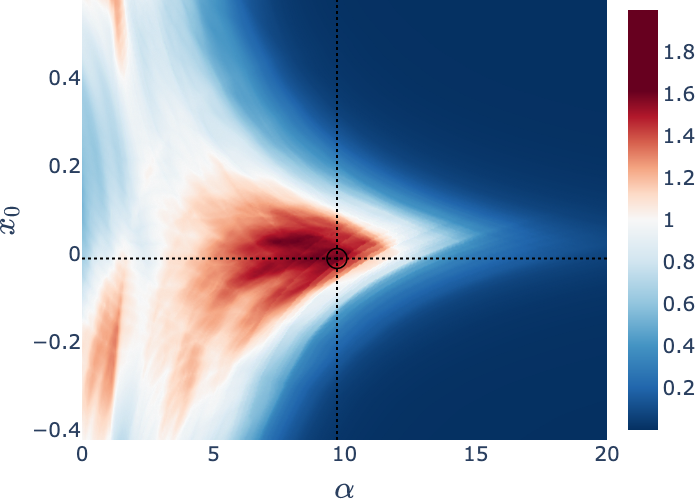} &
        \includegraphics[width=\linewidth,align=c]{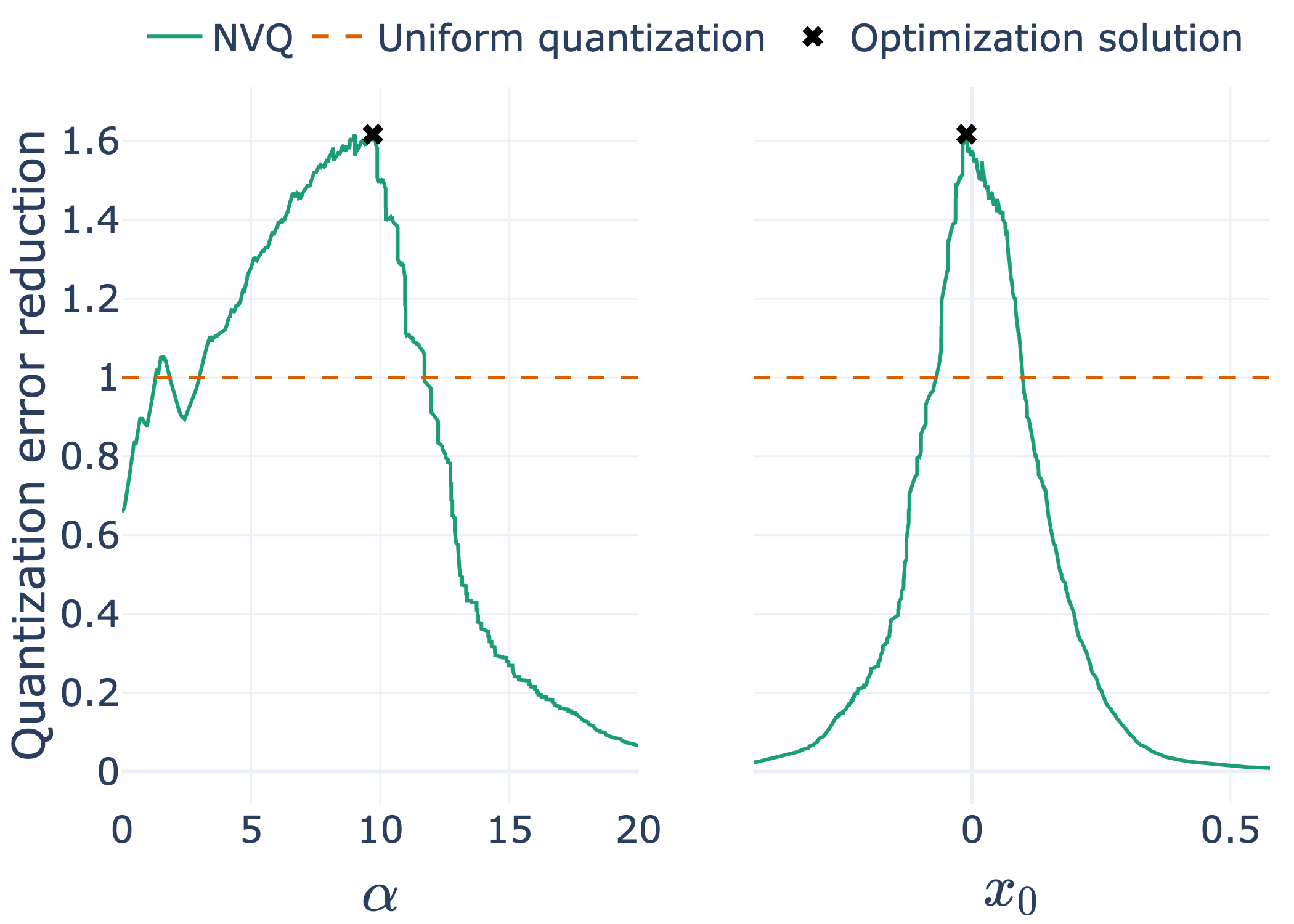} \\
    \end{tblr}
    
    \caption{\zcref{algo:SNES} finds a good maximum of Problem~\zcref[noname]{prob:nuveq} using the nonlinearities presented in \zcref{sec:nonlinearities} for $\beta=4$ bits. \textbf{Left:} the landscape of the objective function in Problem~\zcref[noname]{prob:nuveq}, which is the ratio of the MSE improvement over the uniform quantization (a value of 1 means parity, higher is better), as a function of the nonlinearity parameters for one vector from ada002-100k (see \zcref{tab:datasets}). The solution found by \zcref{algo:SNES} is marked by a black circle. \textbf{Right:} two cross cuts taken at the values corresponding to the found solution.}
    \label{fig:single_vector_ada002-100k-4bits}
\end{figure}

\begin{figure}[t]
    \centering
    \begin{tblr}{
        colspec = {X[c,r,1]X[c,h,10]X[c,h,10]},
    }
        \begin{sideways}
            \small Kumaraswamy
        \end{sideways} &
        \includegraphics[width=\linewidth,align=c]{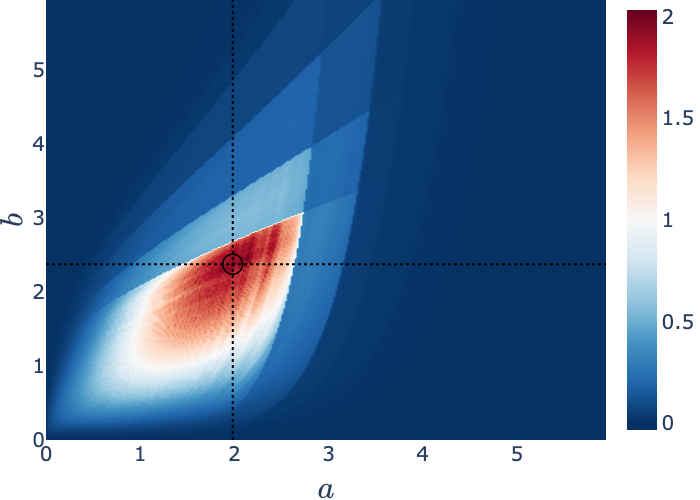} &
        \includegraphics[width=\linewidth,align=c]{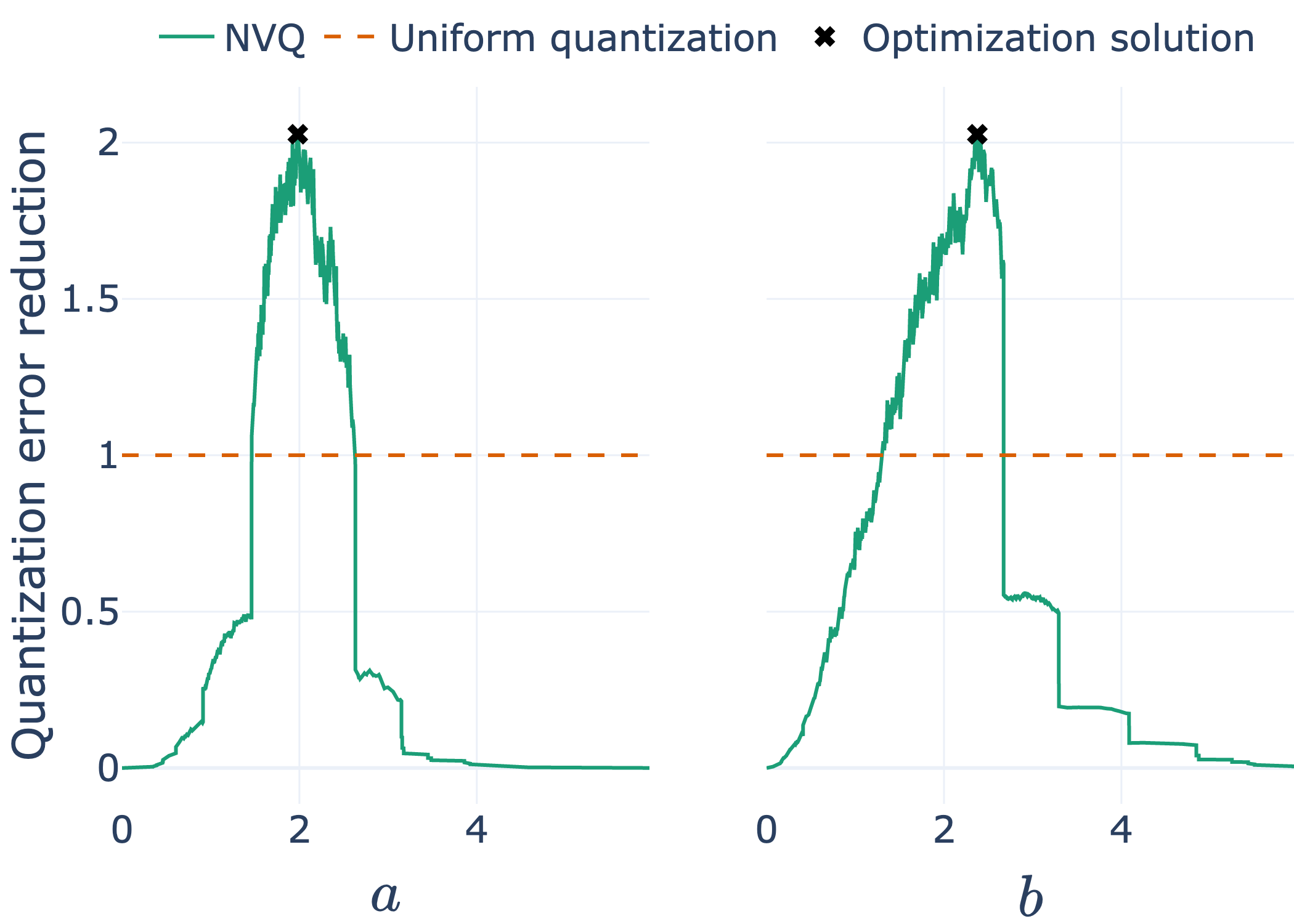} \\

        \begin{sideways}
            \small Log-Log
        \end{sideways} &
        \includegraphics[width=\linewidth,align=c]{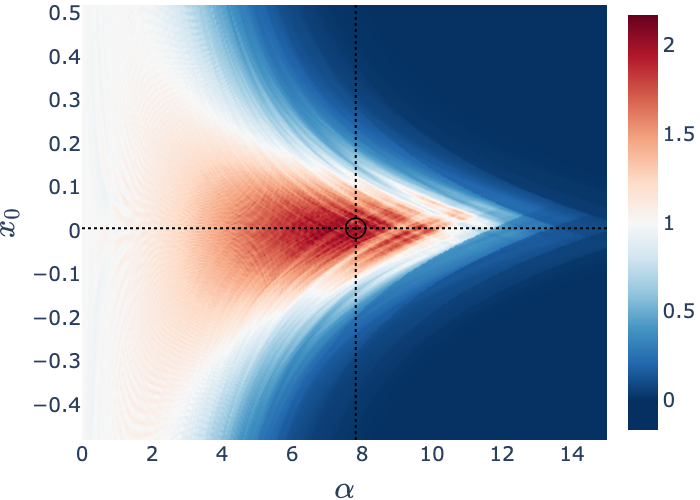} &
        \includegraphics[width=\linewidth,align=c]{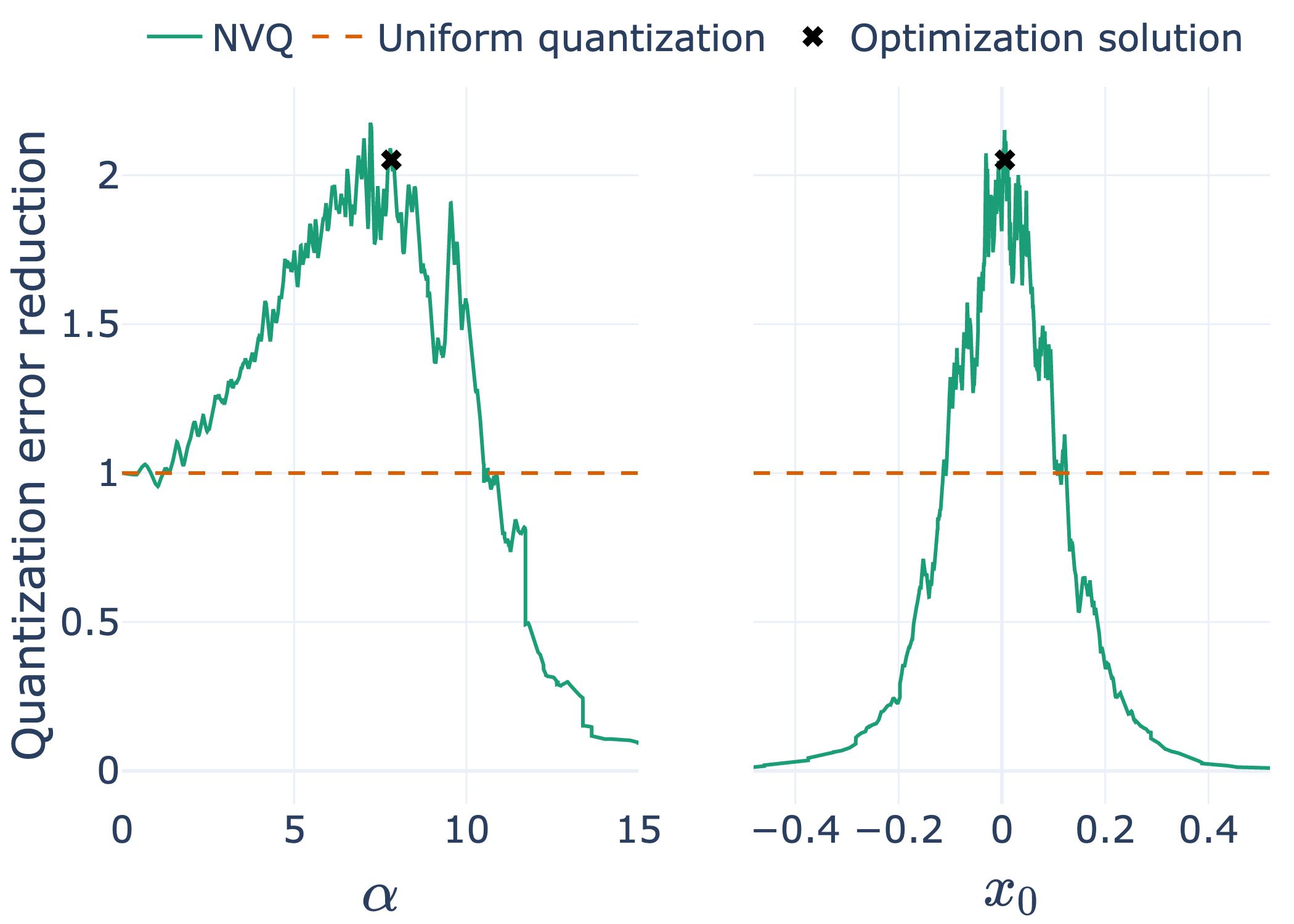} \\

        \begin{sideways}
            \small NQT
        \end{sideways} &
        \includegraphics[width=\linewidth,align=c]{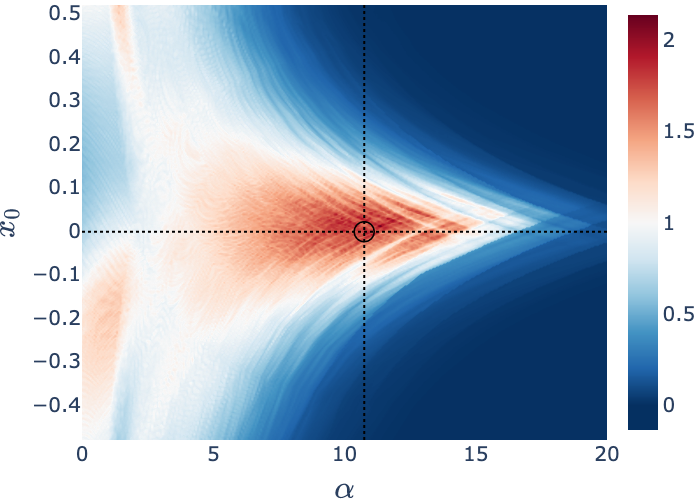} &
        \includegraphics[width=\linewidth,align=c]{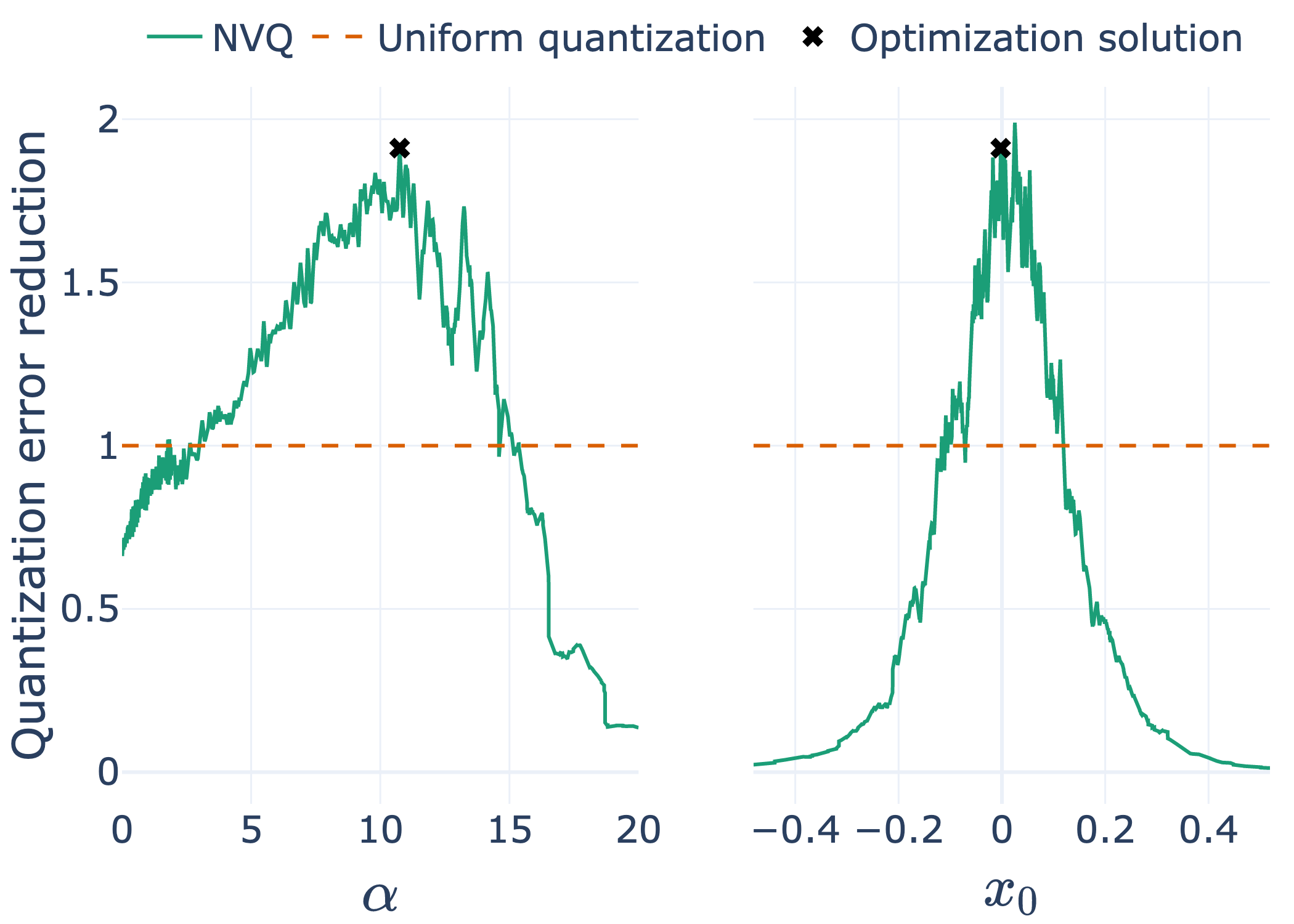} \\
    \end{tblr}
    
    \caption{\zcref{algo:SNES} finds a good maximum of Problem~\zcref[noname]{prob:nuveq} using the nonlinearities presented in \zcref{sec:nonlinearities} for $\beta=8$ bits. \textbf{Left:} the landscape of the objective function in Problem~\zcref[noname]{prob:nuveq}, which is the ratio of the MSE improvement over the uniform quantization (a value of 1 means parity, higher is better), as a function of the nonlinearity parameters for one vector from openai-v3-100k . The solution found by \zcref{algo:SNES} is marked by a black circle. \textbf{Right:} two cross cuts taken at the values corresponding to the found solution.}
    \label{fig:single_vector_openai-8bits}
\end{figure}

\begin{figure}[t]
    \centering
    \begin{tblr}{
        colspec = {X[c,r,1]X[c,h,10]X[c,h,10]},
    }
        \begin{sideways}
            \small Kumaraswamy
        \end{sideways} &
        \includegraphics[width=\linewidth,align=c]{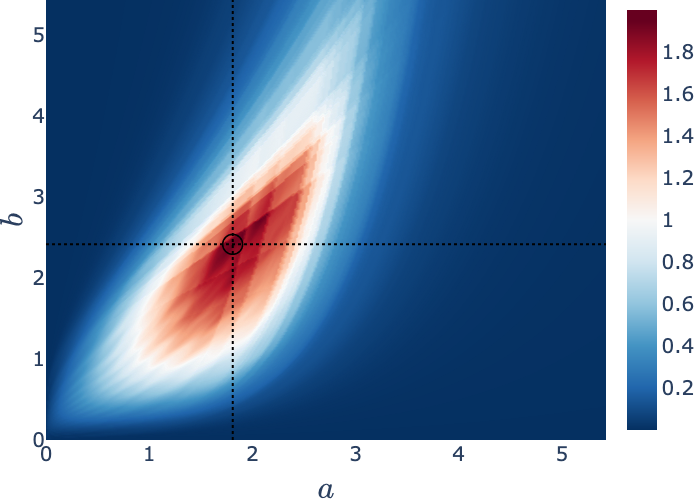} &
        \includegraphics[width=\linewidth,align=c]{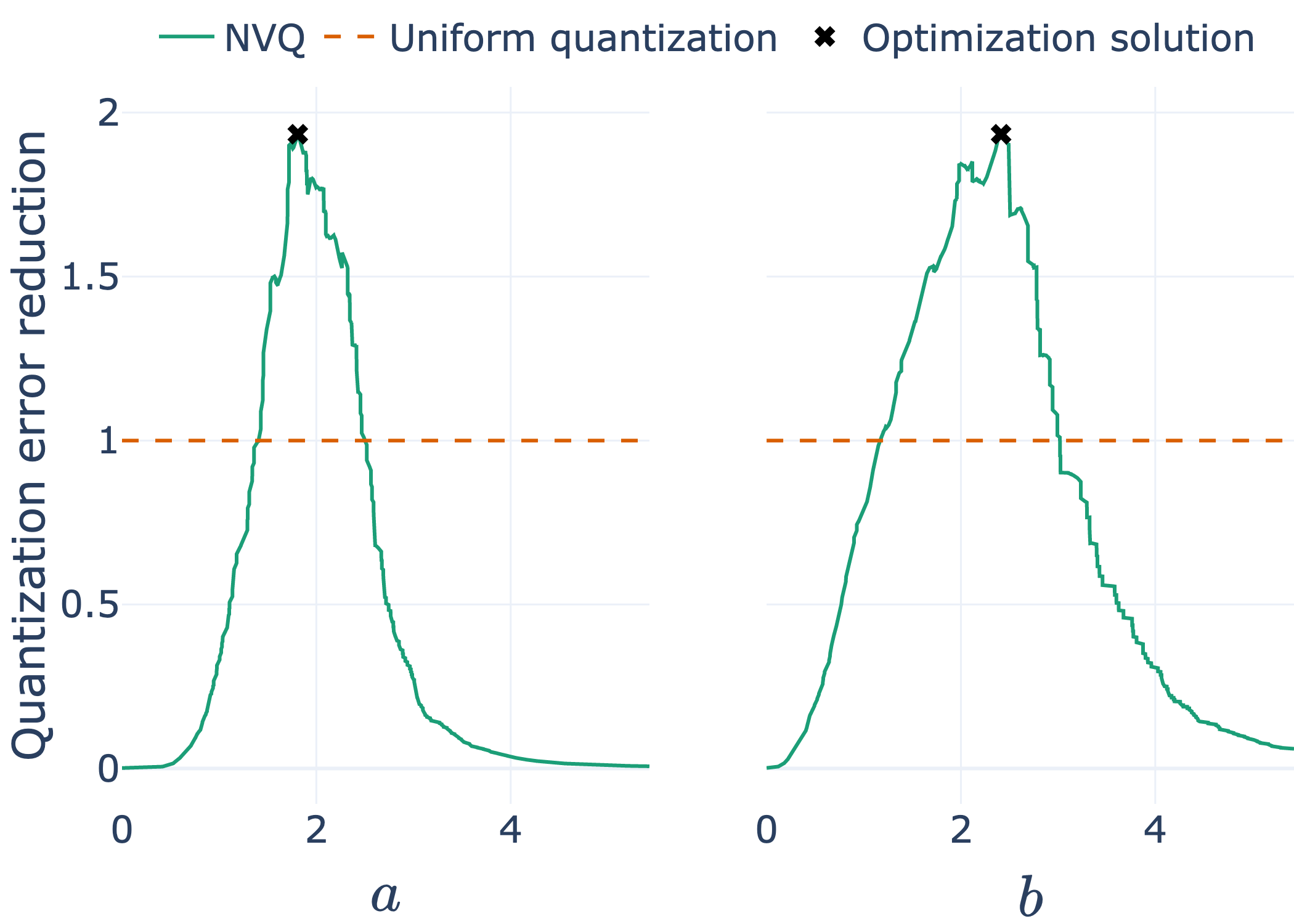} \\

        \begin{sideways}
            \small Log-Log
        \end{sideways} &
        \includegraphics[width=\linewidth,align=c]{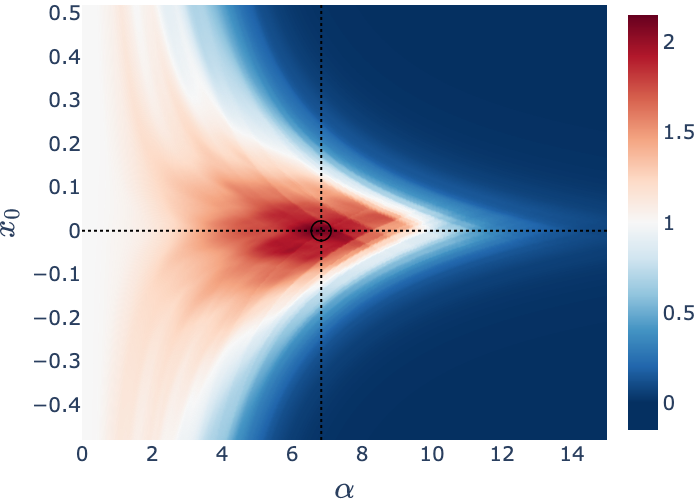} &
        \includegraphics[width=\linewidth,align=c]{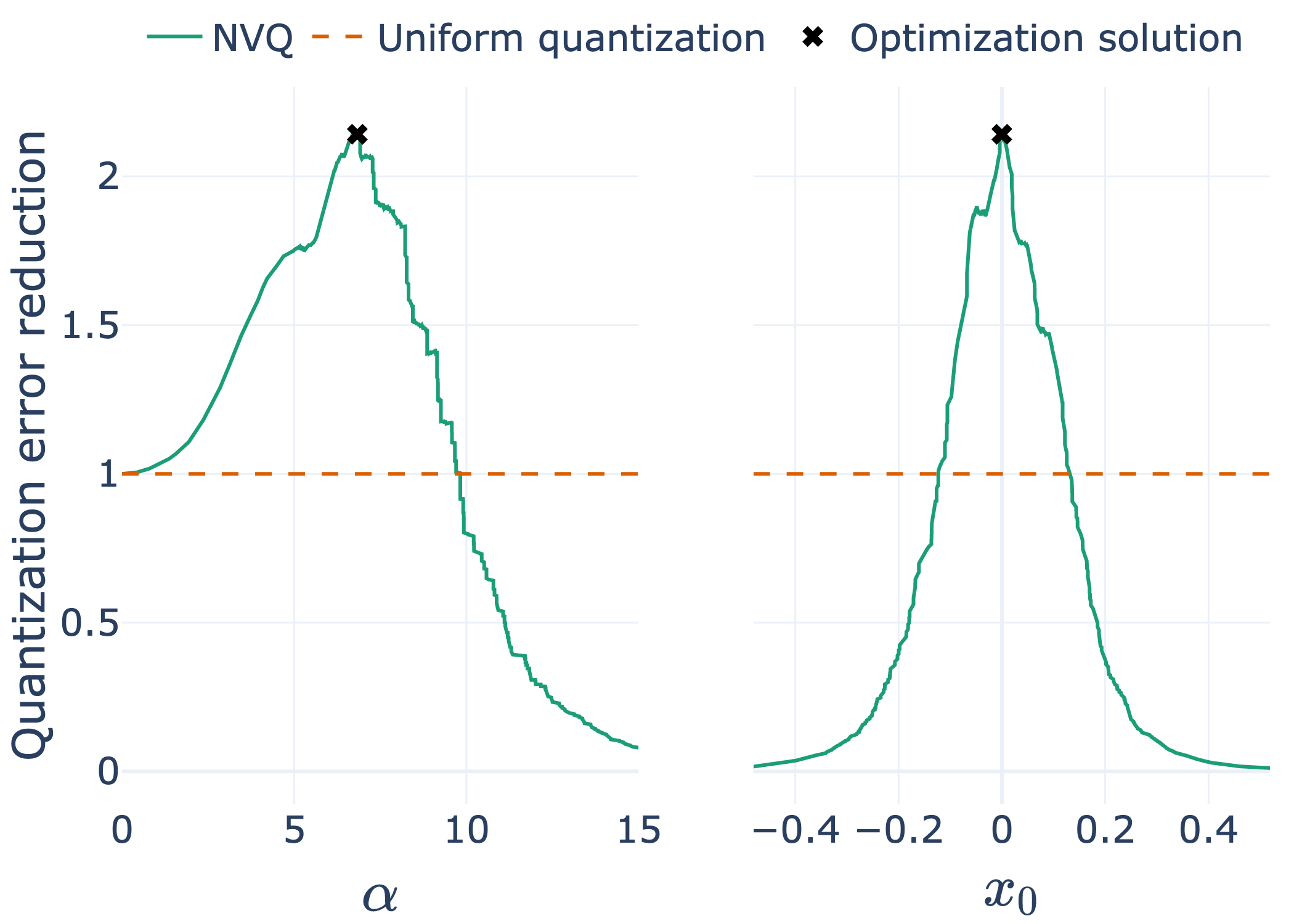} \\

        \begin{sideways}
            \small NQT
        \end{sideways} &
        \includegraphics[width=\linewidth,align=c]{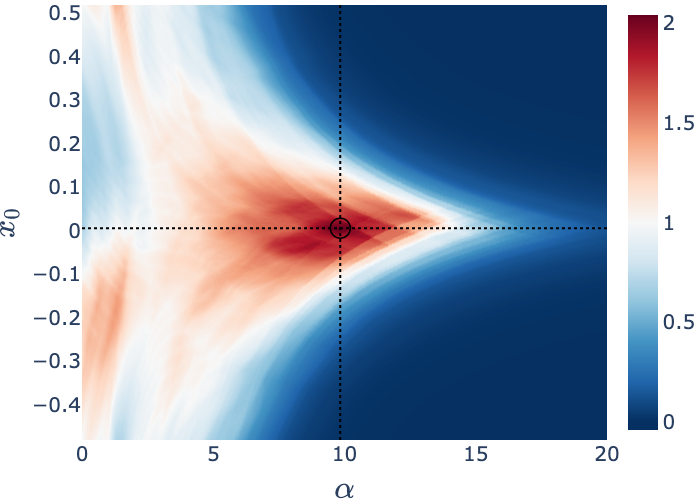} &
        \includegraphics[width=\linewidth,align=c]{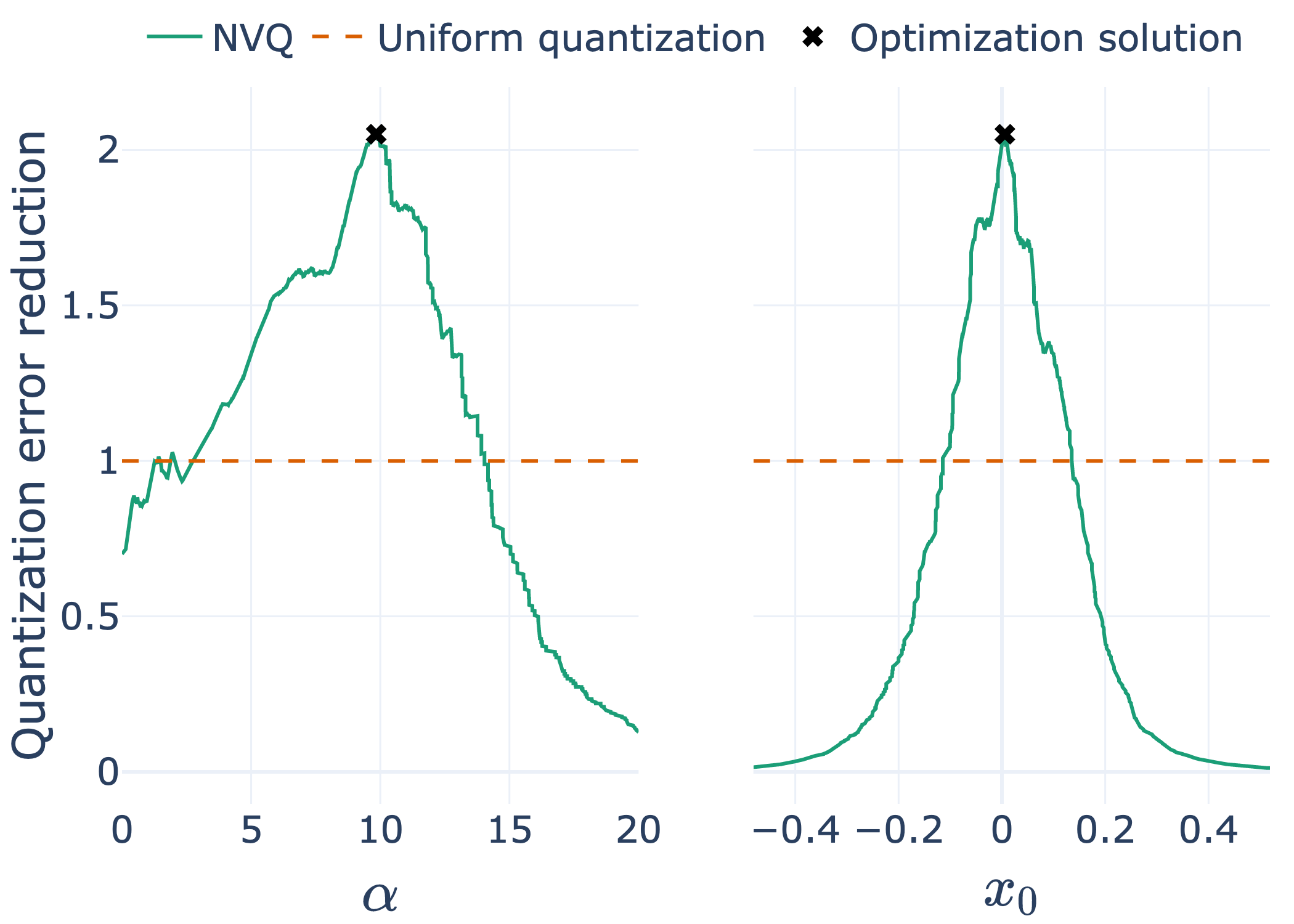} \\
    \end{tblr}
    
    \caption{\zcref{algo:SNES} finds a good maximum of Problem~\zcref[noname]{prob:nuveq} using the nonlinearities presented in \zcref{sec:nonlinearities} for $\beta=4$ bits. \textbf{Left:} the landscape of the objective function in Problem~\zcref[noname]{prob:nuveq}, which is the ratio of the MSE improvement over the uniform quantization (a value of 1 means parity, higher is better), as a function of the nonlinearity parameters for one vector from openai-v3-100k . The solution found by \zcref{algo:SNES} is marked by a black circle. \textbf{Right:} two cross cuts taken at the values corresponding to the found solution.}
    \label{fig:single_vector_openai-4bits}
\end{figure}

\begin{figure}
    \centering
    \includegraphics[width=0.32\linewidth]{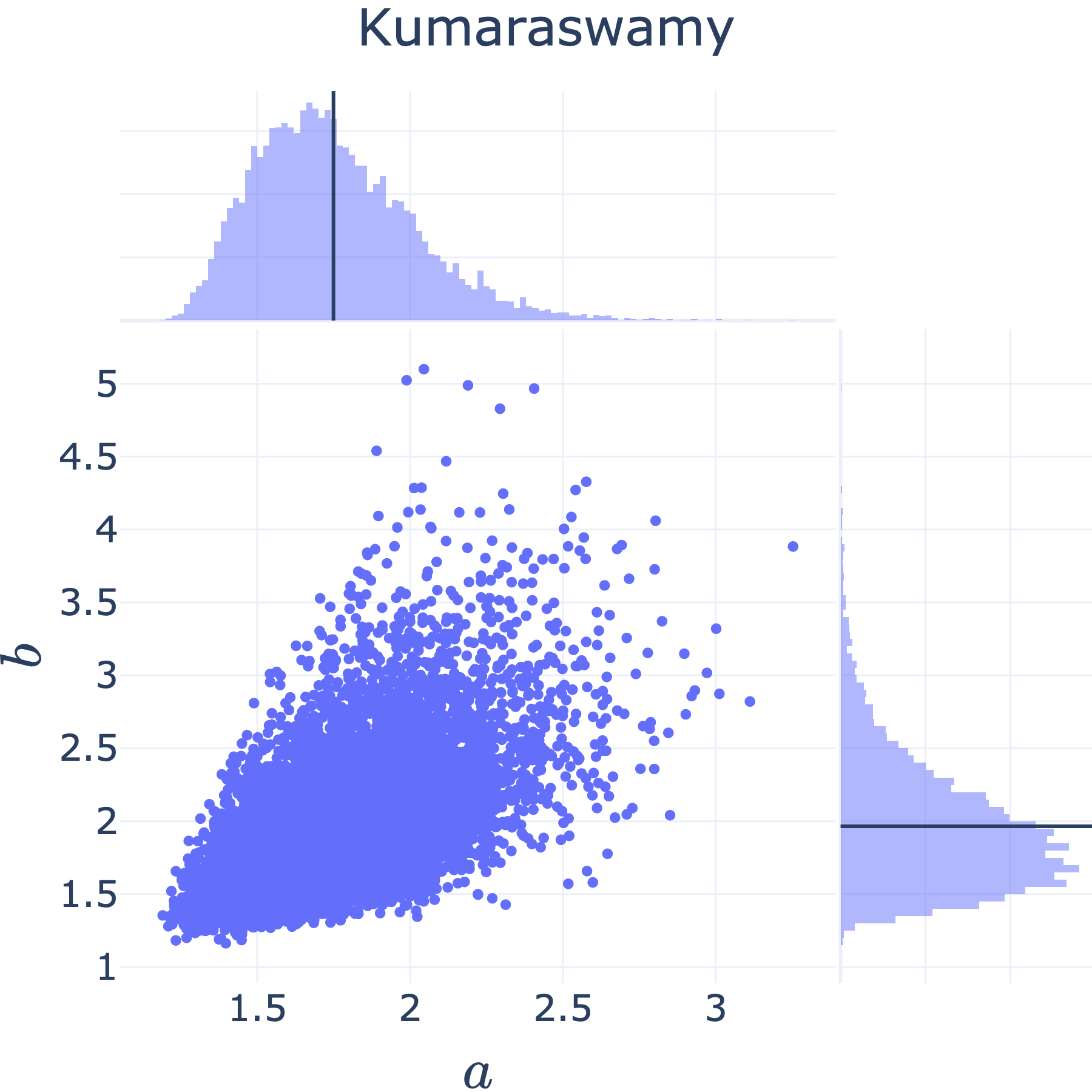}%
    \hfill%
    \includegraphics[width=0.32\linewidth]{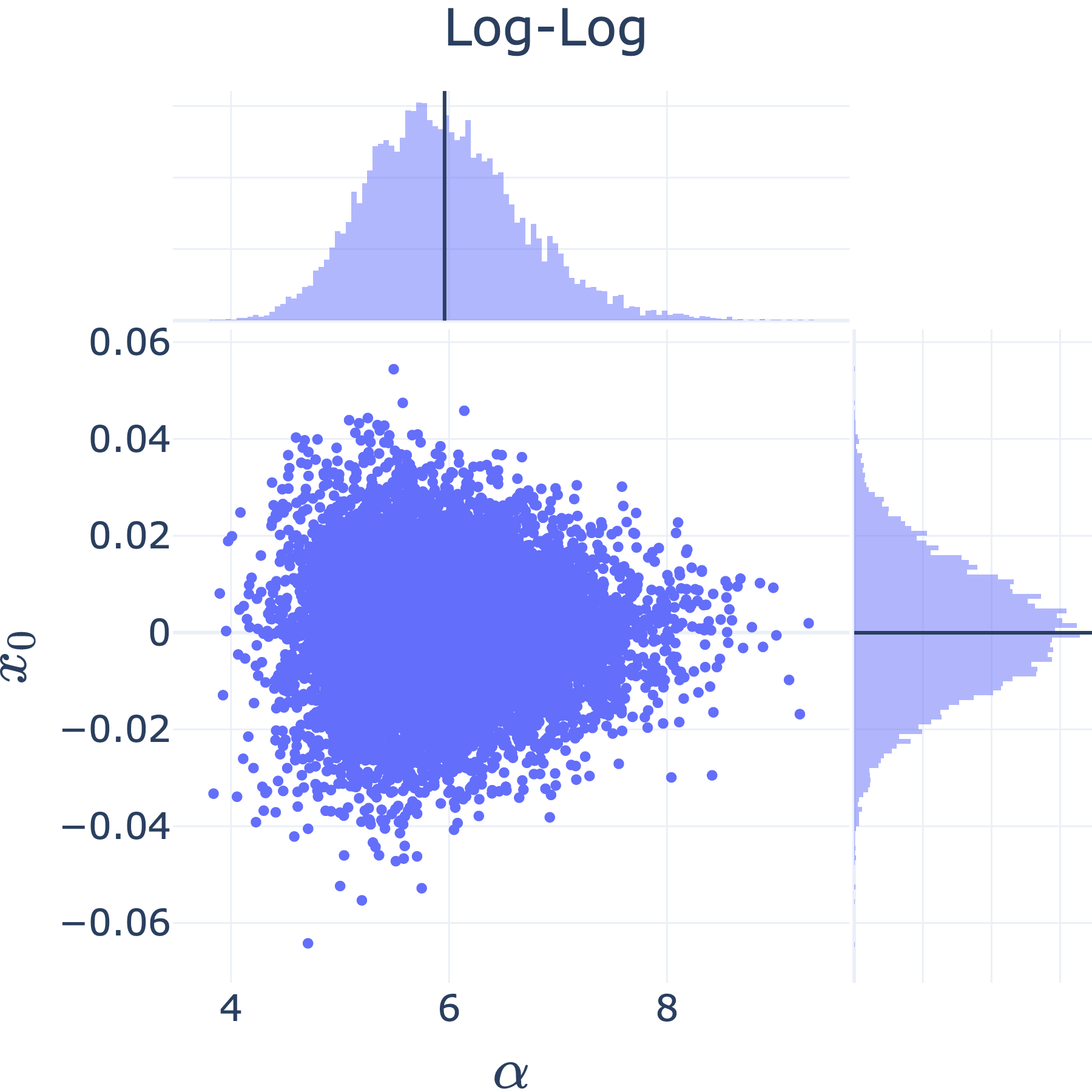}%
    \hfill%
    \includegraphics[width=0.32\linewidth]{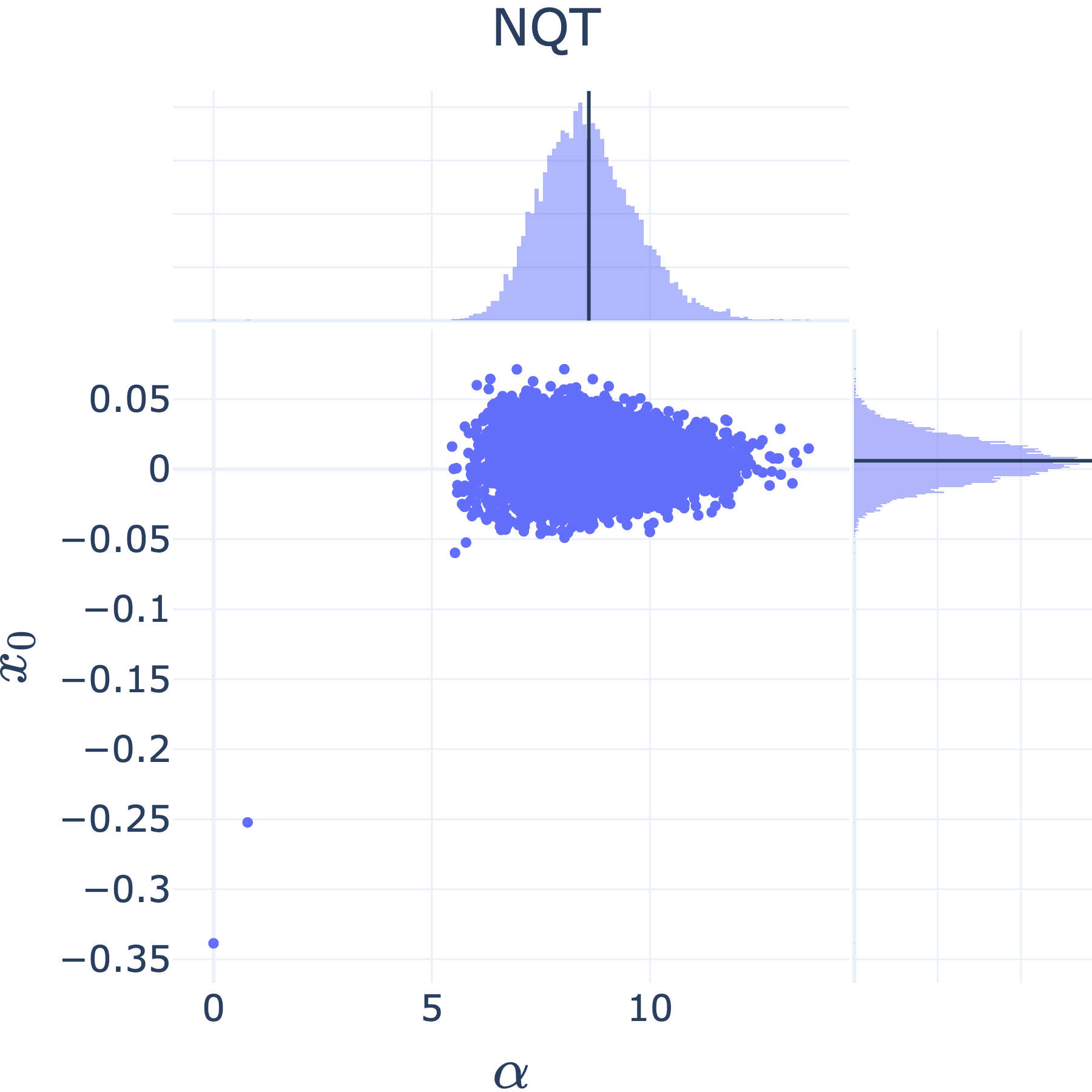}%
    
    \caption{The solutions for $10^4$ vectors from gecko-100k. Different nonlinearity parameters are chosen for each vector.}
    \label{fig:solutions_distribution_gecko}
\end{figure}

\begin{figure}[t]
    \centering
    \includegraphics[width=0.9\linewidth]{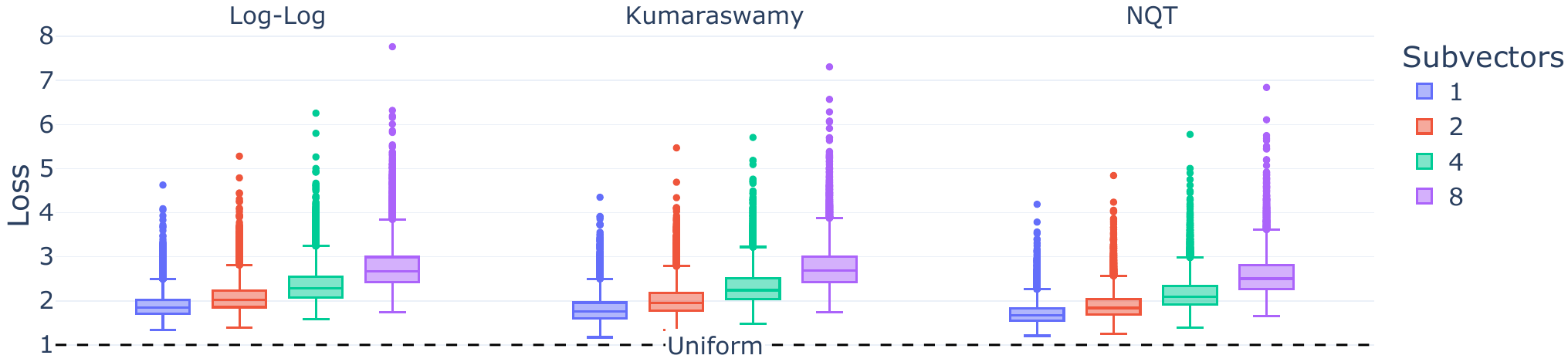}
    
    \caption{Increasing the number of subvectors (depicted here for $\beta=8$) increases the objective function (higher is better) in Problem~\zcref[noname]{prob:nuveq} for $10^4$ vectors from ada002-100k.
    }
    \label{fig:subvectors_loss_continued}
\end{figure}

\begin{figure}[t]
    \centering
    \small
    \begin{tblr}{
        colspec = {X[c,r,1]X[c,r,1]X[c,h,30]},
    }
        \begin{sideways}
            ada002-100k
        \end{sideways} &
        \begin{sideways}
            $\beta=8$ bits
        \end{sideways} &
        \includegraphics[width=\linewidth,align=c]{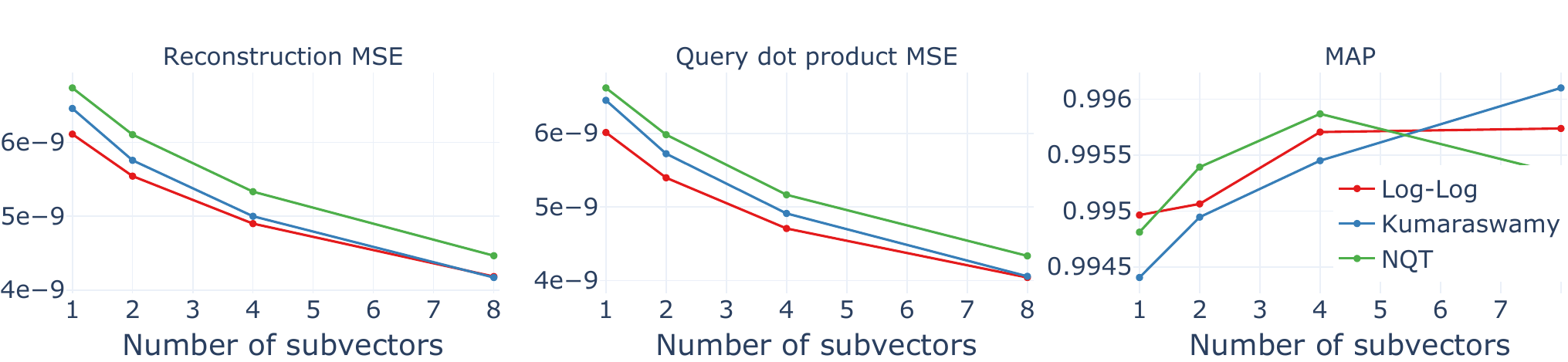}
        \\
        \begin{sideways}
            openai-v3-100k
        \end{sideways} &
        \begin{sideways}
            $\beta=4$ bits
        \end{sideways} &
        \includegraphics[width=\linewidth,align=c]{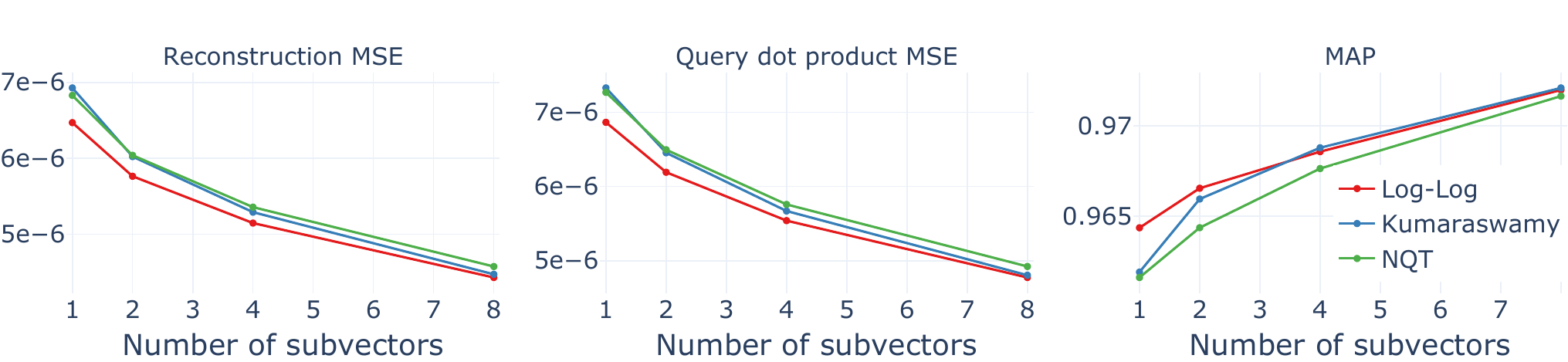}
        \\
        \begin{sideways}
            openai-v3-100k
        \end{sideways} &
        \begin{sideways}
            $\beta=8$ bits
        \end{sideways} &
        \includegraphics[width=\linewidth,align=c]{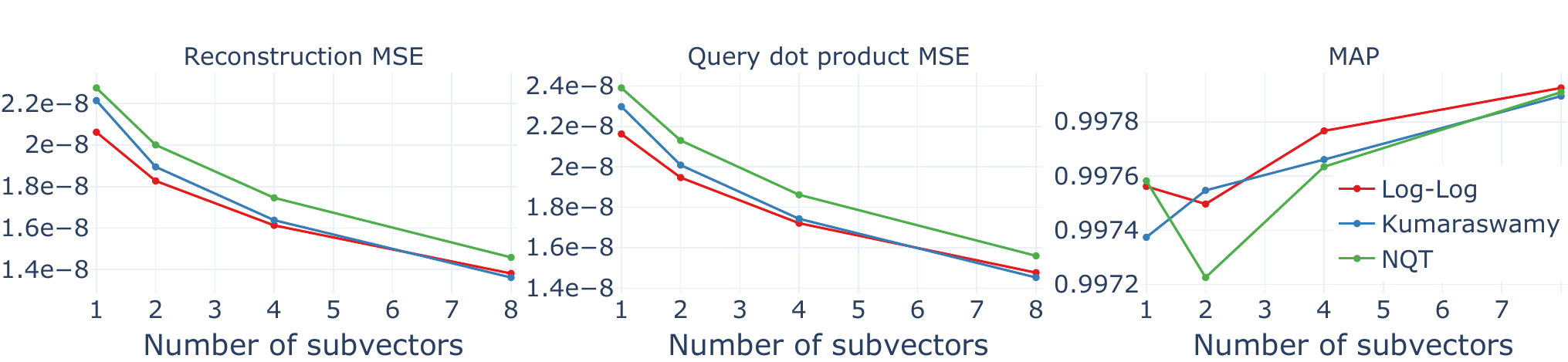}
        \\
    \end{tblr}
    
    \caption{Increasing the number of subvectors lowers the average reconstruction error and the query dot product error, defined as $\sum_{\vect{x} \in \set{X}}(\langle \vect{q}, \vect{x} \rangle - \langle \vect{q}, \widetilde{\vect{x}} \rangle )^2$, and increases the mean average precision (MAP) in openai-v3-100k.}
    \label{fig:subvectors_recall_continued}
\end{figure}

\end{document}